\documentclass[journal]{IEEEtran}

\usepackage{xcolor,soul,framed} %,caption

\colorlet{shadecolor}{yellow}
\usepackage[pdftex]{graphicx}
\graphicspath{{../pdf/}{../jpeg/}}
\DeclareGraphicsExtensions{.pdf,.jpeg,.png}

\usepackage[cmex10]{amsmath}
\usepackage{array}
\usepackage{mdwmath}
\usepackage{mdwtab}
\usepackage{eqparbox}
\usepackage{url}
\usepackage{algorithm}
\usepackage{algorithmic}
\usepackage{amssymb}
\graphicspath{{./figures/}}
\usepackage{floatflt}
\usepackage{multirow}
\usepackage{lipsum}
\usepackage{graphicx,subfigure}
\usepackage{xcolor}

\title{Machine Unlearning Through Fine-Grained Model Parameters Perturbation}

\author{
    \IEEEauthorblockN{Zhiwei Zuo\IEEEauthorrefmark{2}\IEEEauthorrefmark{3}, Zhuo Tang\IEEEauthorrefmark{2}\IEEEauthorrefmark{4}\IEEEauthorrefmark{1}, Kenli Li\IEEEauthorrefmark{2}, Anwitaman Datta\IEEEauthorrefmark{3}\IEEEauthorrefmark{5}}\\
    \IEEEauthorblockA{\IEEEauthorrefmark{2}College of Computer Science and Electronic Engineering, Hunan University, China}\\
    \IEEEauthorblockA{\IEEEauthorrefmark{3}College of Computing and Data Science, Nanyang Technological University, Singapore}\\
    \IEEEauthorblockA{\IEEEauthorrefmark{4}Shenzhen Research Institute, Hunan University, China}\\
    \IEEEauthorblockA{\IEEEauthorrefmark{5}Faculty of Computing, Engineering \& Media, De Montfort University, Leicester, United Kingdom}
    \thanks{Corresponding author: Zhuo Tang (ztang@hnu.edu.cn)}
}

\begin{document}

\maketitle

% === ABSTRACT ====================================================================
% =================================================================================
\begin{abstract}
%\boldmath
Machine unlearning involves retracting data records and reducing their influence on trained models, aiding user privacy protection, at a significant computational cost potentially. Weight perturbation-based unlearning is common but typically modifies parameters globally. We propose fine-grained Top-K and Random-k parameters perturbed inexact machine unlearning that address the privacy needs while keeping the computational costs tractable.

However, commonly used training data are independent and identically distributed, for inexact machine unlearning, current metrics are inadequate in quantifying unlearning degree that occurs after unlearning. To address this quantification issue, we introduce SPD-GAN, which subtly perturbs data distribution targeted for unlearning. Then, we evaluate unlearning degree by measuring the performance difference of the models on the perturbed unlearning data before and after unlearning. Furthermore, to demonstrate efficacy, we tackle the challenge of evaluating machine unlearning by assessing model generalization across unlearning and remaining data. To better assess the unlearning effect and model generalization, we propose novel metrics, namely, the forgetting rate and memory retention rate. By implementing these innovative techniques and metrics, we achieve computationally efficacious privacy protection in machine learning applications without significant sacrifice of model performance. A by-product of our work is a novel method for evaluating and quantifying unlearning degree.

\end{abstract}

% === KEYWORDS ====================================================================
% =================================================================================

\IEEEpeerreviewmaketitle

% ====================================================================
% ====================================================================
% ====================================================================

% === I. INTRODUCTION =============================================================
% =================================================================================
\section{Introduction}

\IEEEPARstart{U}{sers} may want the removal of personal data in possession of an organization, typically deleting it randomly based on their specific needs. In several jurisdictions, this has compliance implications under regulations such as the General Data Protection Regulation (GDPR) in the European Union \cite{goddard2017eu} and the California Consumer Privacy Act in the United States  \cite{goldman2020introduction}. Machine learning models trained using such data retain their influence, thus risking privacy compromises \cite{wu2023privacy}. For instance, in a recommendation system using collaborative filtering \cite{ko2022survey,liu2023pre,rajput2023recommender}, if the model does not eliminate the influence of removed data (which we call `unlearning data'), it may continue to make recommendations based on user similarity.

Separately, one may want to remove the influence of a subset of data from a model trained with a corpus of data, possibly because the information in the subset corpus is obsolete, or even wrong e.g., determined to be fake/misinformation subsequent to its original use in model training.

The task to address this concern by eliminating the influence of unlearning data on trained models has thus emerged in the recent years as the budding topic of machine unlearning.

Retraining from scratch, due to its extensive time and computational resource consumption, has become impractical, thus making  machine unlearning extremely necessary. Common unlearning strategies can be divided into two catagories: exact unlearning and inexact unlearning.

Exact machine unlearning methods \cite{thudi2022necessity, bourtoule2021machine} are designed to completely eliminate the influence of unlearning data on a model. Although some methods explore the impact of data on models, the Influence Function provides insights into such impacts, but it has limitations. The non-convexity of the loss functions used in deep learning models leads to significant approximation errors, and the computational burden of calculating the inverse of the Hessian matrix in the Influence Matrix limits its application in deep learning environments \cite{basu2020influence, li2023selective}. In contrast, the Single-Step Sample Erasure (SSSE) \cite{peste2021ssse} utilizes the Fisher Information Matrix to sidestep computational hurdles, achieving a more precise approximation.

Differently, inexact machine unlearning \cite{DBLP:journals/corr/abs-2201-06640,suriyakumar2022algorithms} aims to reduce time and resources requirements by selectively eliminating the data's influence on the model. Current inexact machine unlearning methods based on model weight perturbation typically involve adding random Gaussian noise to all model parameters and then training for several epochs based on remaining data. We propose to achieve inexact unlearning using a more fine-grained perturbation. To that end, we design two inexact machine unlearning strategies: Random-k and Top-K. Here, Random-k represents perturbing k\% of the parameters, while Top-K represents perturbing the top K parameters. To avoid notational confusion, we distinguish the two approaches by using lower and upper case respectively. By selectively perturbing a small subset of parameters, we aim to achieve the desired effect of machine unlearning efficiently while also reducing the impact on the overall model performance. 

Even if we conduct machine unlearning successfully, evaluating unlearning effectiveness remains a challenge, especially in terms of how to quantify the degree of unlearning. This is because when different algorithms learn from the same data set, they may acquire very similar knowledge and features. Consequently, when faced with independent and identically distributed (i.i.d) datasets, even if an unlearning approach has eliminated the influence of unlearning data, the model may still show approximate accuracy on the unlearning data to the remaining data due to the model's generalization property \cite{nakkiran2020distributional}. This makes it challenging to evaluate the effectiveness of the unlearning process.

In practice, to measure the effectiveness of unlearning, current metrics can be divided into two aspects. The first is general performance evaluation, which includes assessing the accuracy \cite{marchant2022hard} of unlearning model predictions, unlearn time corresponding to the unlearning request \cite{bourtoule2021machine}, relearn time for reaching the accuracy of source model \cite{tarun2023fast}, membership inference attack (MIA) \cite{shokri2017membership,hu2022membership}. The second focuses on model indistinguishability \cite{golatkar2020eternal, kurmanji2023towards}, completeness, activation distance \cite{wu2020deltagrad}, Jensen–Shannon (JS)-Divergence \cite{chundawat2023zero}, and other metrics that quantify the differences between the unlearning and retraining models.

\begin{figure}
\centering
  \includegraphics[width=0.4\textwidth]{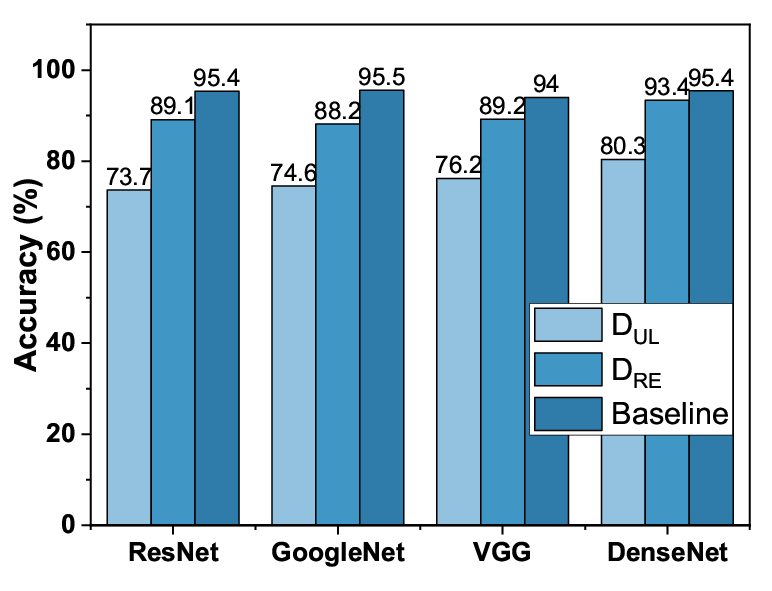}
  \caption{Random-k unlearning on CIFAR-10, accuracy of $D_{RE}$ and $D_{UL}$ after training for 50 epochs. Baseline is the test accuracy on the source model.}
  \label{fig:iid}
\end{figure}

However, the existing metrics are unable to reveal the extent of unlearning. Suppose splitting training data into two parts: unlearning data $D_{UL}$ and remaining data $D_{RE}$, where $D = D_{UL} \cup D_{RE}$, $|D_{UL}| \ll |D_{RE}|$. As shown in Fig.\ref{fig:iid}, which shows an early glimpse of some results achieved by one of our own unlearning techniques), while the accuracy may vary across $D_{UL}$ and $D_{RE}$ for each model, it consistently maintains a high accuracy for $D_{UL}$. Consequently, it becomes challenging to quantify the extent to which the model eliminates the influence of $D_{UL}$. Since the performance difference of unlearning model $M_{UL}$ on $D_{UL}$ and $D_{RE}$ is relatively small. We attribute this to two reasons, first, a good generalization capability of the model, second, the characteristic that the training data followed an independent and identically distributed (i.i.d) property. 

In order to amplify the performance difference between the unlearning model $M_{UL}$ on $D_{UL}$ and $D_{RE}$, we propose a SPD-GAN (Slightly Perturbed Distribution-Generative Adversarial Network) to break the i.i.d property of the unlearning data. For the unlearning data $D_{UL}$, we introduce perturbations to it's distribution, thus we gain perturbed unlearning data $D_p$. We limit the perturbations to be very slight so that the accuracy of source model $M$ on $D_p$ is close to the accuracy of $M$ on $D_{UL}$. Suppose $P(M, D)$ represents the performance of model $M$ on data $D$, seeing in the Eq.\ref{equation1}, if the perturbed unlearning data $D_p$ exhibits significant performance difference between source model and unlearning model, we can then obtain the unlearning degree of each unlearning strategies.

\begin{align}
    \centering
    & P(M, D_p) \gg P(M_{UL}, D_p) \nonumber \\
    & s.t. ~ P(M, D_p) \approx P(M, D_{UL}) 
    \label{equation1}
\end{align}

In addition to measuring the degree of unlearning, in the fine-grained inexact machine unlearning strategies we employ, we also consider both the impact of model unlearning and the generalization performance of the model. A metric called \emph{forgetting rate} is designed to assess the effect of model unlearning, while \emph{memory retention rate} and \emph{similarity} are employed to evaluate the model's indistinguishability. 

Our main contributions are:

\begin{itemize}
    \item  We adopt two fine-grained inexact machine unlearning strategies of Random-k and Top-K to quickly achieve the unlearning effect with minimal model parameters perturbed instead of perturbed globally while ensuring its generalization properties (Section \ref{approach}).
    
    \item We design a novel unlearning degree evaluation method by proposing a SPD-GAN to slightly perturb the unlearning data, using the performance difference of the perturbed unlearning data on the source model and unlearning model to approximate the degree of unlearning (Section \ref{unlearning effectiveness}). Experimental results show that our proposed Top-K can achieve the deepest degree of unlearning in comparison to other inexact unlearning strategies (as discussed in Section \ref{unlearning degree evaluation}).
    
    \item We design measurement metrics \emph{forgetting rate}, \emph{memory retention rate} and \emph{similarity} to assess unlearning effectiveness and the unlearning model's generalization properties. We carry out a theoretical analysis in Section \ref{complexity Analysis} of the acceleration of inexact machine unlearning strategies, and represent it as \emph{acceleration ratio} in our experiments (as discussed in Section \ref{effectiveness of Top-K and Random-k}).
\end{itemize}

In the subsequent sections, we first explore the related work in Section \ref{related work}, then describe the methodology in Section \ref{methodology}, report the findings from our experiments in Section \ref{experiments}, and finally, draw conclusions and discuss future research directions in Section \ref{conclusion}.

\section{Related Works}\label{related work}
\subsection{User Privacy}
For data-driven decision-making systems, the data used is primarily collected from the internet or exposed by users when they interact with such systems, much of which is highly sensitive and confidential. There are many methods trying to serve the function of privacy protection in various scenarios. Traditional approaches achieve decoupling by separating parts of the data that are directly related to the user, perform generalization by erasing or replacing specific details in the data or protect privacy using other anonymization techniques such as data desensitization and data perturbation \cite{gadotti2024anonymization}. However, these methods may not be effective when prior knowledge is unavailable.

Differential privacy (DP) \cite{jain2023price} provides a strict definition of privacy, making it the de facto standard across various data types in the field of privacy protection. Differential privacy (DP) operates by adding well-calibrated noise to individual data points or database queries, ensuring that the results of data analysis do not significantly change due to the presence or absence of any single data point. Local differential privacy (LDP) \cite{yang2023local} perturbs data before it leaves a user's local device, ensuring that only the data owner retains access to the original data. However, LDP can introduce excessive noise into the dataset. Other approaches, such as dPA \cite{kocher2011introduction} add perturbation to the objective function, while PATE \cite{papernot2018scalable} uses knowledge aggregation and transmission. However, the process of adding noise in DP is irreversible, making it unsuitable for machine unlearning tasks since the protection cannot be removed once data is protected by noise, and more crucially in our context, it also is not suitable for removing influence of data from a model that has already been trained with a given data.

Federated learning (FL) is a decentralized approach to protect user privacy, where training data remains distributed across multiple devices and servers, and the model is trained without centralizing the data \cite{wei2020federated}. FL methods based on differential privacy have been proposed to protect user privacy, with the principal focus on attacks and defenses \cite{xiao2022sbpa} in FL systems. Furthermore, recent advancements in FL incorporate various cryptographic techniques, such as secure multi-party computation and homomorphic encryption, to enhance the security and privacy of the training process. This ongoing research strengthens the privacy-preserving capabilities of FL.

Other decentralized privacy protection technologies, like blockchain \cite{reijsbergen2023piechain}, distribute data across many different nodes in the network to reduce the risk of data tampering or unauthorized access, while also using encryption technologies to protect the transmission and access of data.

\subsection{Machine Unlearning}
Several approaches have emerged to address the limitations of retraining models from scratch for machine unlearning, since it is generally impractical. the most common approach is SISA (Sharded, Isolated, Sliced, and Aggregated) training \cite{bourtoule2021machine}, which splits data into shards and trains them in isolation for speed and effectiveness when unlearning. Liu et al. \cite{liu2022continual} combine continual learning with private learning inspired by the SISA approach, but SISA-based methods often consume a lot of storage. Following the combination of machine unlearing and continual learning, Zuo et al. \cite{zuo2024ecil} propose a embedding-based framework, which modifies the data rather than the model itself to achieve unlearning goal by using vector databases.

Other approaches focus on evaluating the impact of specific data points or sub-datasets on trained models. Influence function \cite{koh2017understanding} approximates the parameter change when data is up-weighted by a small value, but its effectiveness in deep learning is limited by non-convex loss functions and Taylor's approximation errors \cite{basu2020influence}. Peste et al. propose a Single-Step Sample Erasure (SSSE) using the Fisher Information Matrix (FIM) to avoid expensive computation and Hessian inversion \cite{peste2021ssse}. Some works use influence functions in Graph Neural Networks (GNNs) for unlearning, but these models are relatively shallow, and Hessian inversion is more accurate, allowing for exact  unlearning.

In addition to combining with continual learning, machine unlearning is also integrated with other approaches, such as federated learning, forming a research area known as federated unlearning \cite{liu2021federaser} \cite{zhao2023federated}. In federated unlearning, considerations are made for clients leaving the federation or for unlearning data corresponding to clients. For instance, Liu et al. \cite{liu2021federaser} propose FedEraser, which unlearns a client's data by reconstructing an unlearned global model instead of retraining from scratch. Zhao et al. \cite{zhao2023federated} utilize knowledge distillation techniques to recover the contribution of client models to achieve federated unlearning.

While the strategy for unlearning is crucial, the verification of unlearning is also a critical issue. Membership Inference Attack \cite{humembership,ma2022learn} is a common verification method. Utilizing the characteristics of backdoor attacks, users embed a specific backdoor trigger into the data, and then use the backdoor attack to test whether the model has been trained on this data or has successfully eliminated the influence of this data.

\subsection{Adversarial Learning}

Even though neural networks are widely applied, their robustness is hard to ensure. An adversarial sample is the data generated by adding slight perturbation $\delta x$ to raw data $x$, which can cause the neural network to misclassify yet human can still make the correct classification. This leads to potentially fatal dangers in critical areas such as self-driving or biometric authentication, where robust models are essential.

For perturbation generation, Szegedy et al. \cite{szegedy2013intriguing} reveal the vulnerability of neural networks to adversarial examples and introduce the L-BFGS (Limited Memory Broyden-Fletcher-Goldfarb-Shanno) attack method. Due to the high time consumption associated with L-BFGS, Goodfellow et al. \cite{goodfellow2014explaining} propose the fast gradient sign method (FGSM) which optimizes $L_{\infty}$ distance. This method is furthermore refined by Carlini and Wagner into the $C\&W$ method \cite{carlini2017towards}. Additionally, Papernot et al. \cite{papernot2016limitations} propose the Jacobian-based saliency map attack (JSMA), which optimizes under $L_0$ distance. DeepFool \cite{moosavi2016deepfool} looks for the minimal change necessary to deceive the model into misclassifying an input under the specific condition that the perturbation vector is orthogonal to the hyperplane representing the classifier's decision boundary. Generative Adversarial Network (GAN) \cite{ding2023tmg}  is also a promising approach to generate adversarial samples. Each of these methods demonstrates different aspects of the neural networks' vulnerabilities and their respective optimizations highlight various approaches to improving attack efficiency.

\begin{figure*}[!htbp]
    \centering
    \includegraphics[width=0.95\textwidth]{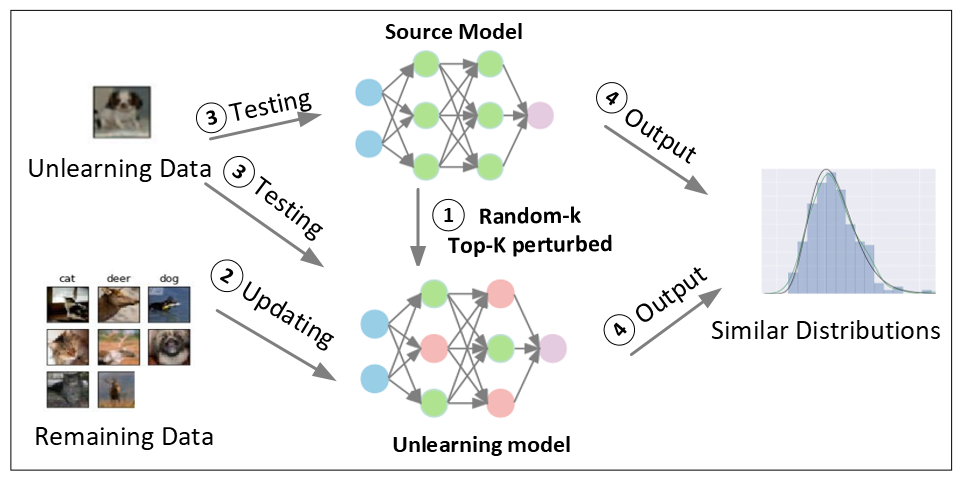}
    \caption{Machine Unlearning process using Random-k or Top-K perturbation strategies. When Random-k/Top-K strategies are applied, partial parameters changed (marked pink on unlearning model).}
    \label{fig:framework}
\end{figure*}

Moreover, these techniques have paved the way for numerous subsequent studies in adversarial machine learning, illustrating the continuous evolution of this field. In other words, leveraging adversarial training enhances the robustness of the model and better identify and defend against malicious attacks. Smoothed ViTs \cite{salman2022certified} are trained using only one columns of original data while the remaining are black. Pyramid Adversarial Training \cite{herrmann2022pyramid} generates adversarial samples and perform a matched Dropout technique and stochastic depth regularization to enhance the robustness.

The concept of perturbation through the addition of noise is also employed in machine unlearning tasks. Neel et al.\cite{neel2021descent} utilizes random Gaussian noise perturbations to the source model $M$'s parameters, constrained by ($\epsilon, \delta$), ensuring that the model remains indistinguishable from any point within a small neighborhood of the optimal model.

\begin{equation}
    \begin{aligned}
    \mathbb{P}(f_m(M(D)) \in S) \leq e^\epsilon \mathbb{P}(f_m(M_{UL}(D_{RE})) \in S) +\delta  \\
    and\quad \mathbb{P}(f_m(M_{UL}(D_{RE})) \in S) \leq e^\epsilon \mathbb{P}(f_m(M(D)) \in S) +\delta 
    \end{aligned}
\end{equation}

Where $f_m$ denotes a random mechanism, with model parameters as its input. Additionally, $S$ represents the subset of the model parameter space. In the status quo, existing adversary-based unlearning methods concentrate on ($\epsilon, \delta$)-perturbations, which involve perturbing all the parameters of the model.

\section{Methodology}
\label{methodology}
\subsection{Problem Definition}

In this paper, we focus on machine unlearning problem. Let $D$ represents the training data set which contains $n$ samples  $\left(x_1, y_1\right), \left(x_2, y_2\right),...,\left(x_n, y_n\right)$. For each sample, $x \in X$ and $y \in Y$. These data are posted by multiple users, collected by institutions or organizations, and then labeled. Suppose $D = D_{UL} \cup D_{RE}$ and $D_{UL} \cap D_{RE} = \emptyset$, here $D_{UL}$ represents the unlearning data, which is absolutely randomly chosen by users to be unlearned and $D_{RE}$ represents the remaining data after unlearning data removal. 

Given a trained model $M$ with parameter $\omega$ on data set $D$, when conducting machine unlearning, our goal is to gain a target model $M_{UL}$ with parameter $\theta$ while the number of perturbed parameters should be as few as possible, and this $M_{UL}$ eliminates the influence of $D_{UL}$ and holds a quite good performance on $D_{RE}$. 
Suppose $\omega'$ is the parameter after fine-grained perturbation. During fine-tuning, the parameters are represented by $\theta$. And as previously mentioned, we assume $P(M,D)$ as the performance of a model $M$ on data $D$, thus, the effect that our fine-grained perturbation machine unlearning aims to is :

\begin{align}
    &P(M_\theta, D_{UL})\ll P(M_\theta, D_{RE}) \nonumber\\  
    &P(M_\theta, D_{RE})\approx P(M_\omega, D_{RE}) \nonumber \\
    & minimize~ \|\omega'-\omega\|_0
\end{align}
where $\|\omega'-\omega\|_0$ is $L_0$ norm, representing the number of perturbed parameters.

We redefine the objective as a loss function, aiming to maximize the prediction error on unlearning data by guiding the model to make predictions more random or close to its untrained state. At the same time, it is necessary to maintain the performance on the remaining data as much as possible.

\begin{equation}
    \mathcal{L}=\mathbb{E}_{x\in D_{UL}}\|\nabla_\theta \ell(f(x),y)\|^2- \mathbb{E}_{x\in D_{RE}}\|\nabla_\theta \ell(f(x),y)\|^2
\end{equation}

Our fine-grained perturbation-based machine unlearning involves two stages: perturbing and fine-tuning. After applying perturbations to the model, the similarity of the output distributions of the model $M$ and $M_{UL}$ guides whether the fine-tuning should be terminated.
Assuming the posterior distribution of $M$ and $M_{UL}$ are $M\left(D_{UL}\right) \sim P_1$ and $M_{UL}\left(D_{UL}\right)\sim P_2$. When $P_1$ and $P_2$ are quite similar, we claim machine unlearning is conducted.

\subsection{Our Unlearning Approaches}
\label{approach}
The parameters in the machine learning model store the high-dimensional features of the data, so in order to eliminate the influence of unlearning data on the model, inspired by adversarial learning \cite{szegedy2013intriguing}, we consider adding slight perturbation to trained model and then perform several fine-tuning epochs. Current machine unlearning methods based on parameters perturbation basically change all the parameters of the model or use tricks such as regularization or Dropout to improve the efficiency of unlearning. In contrast, we aim to perturb the parameters of the model at a finer granularity.
\begin{figure*}[!htbp]
    \centering
    \includegraphics[width=0.95\textwidth]{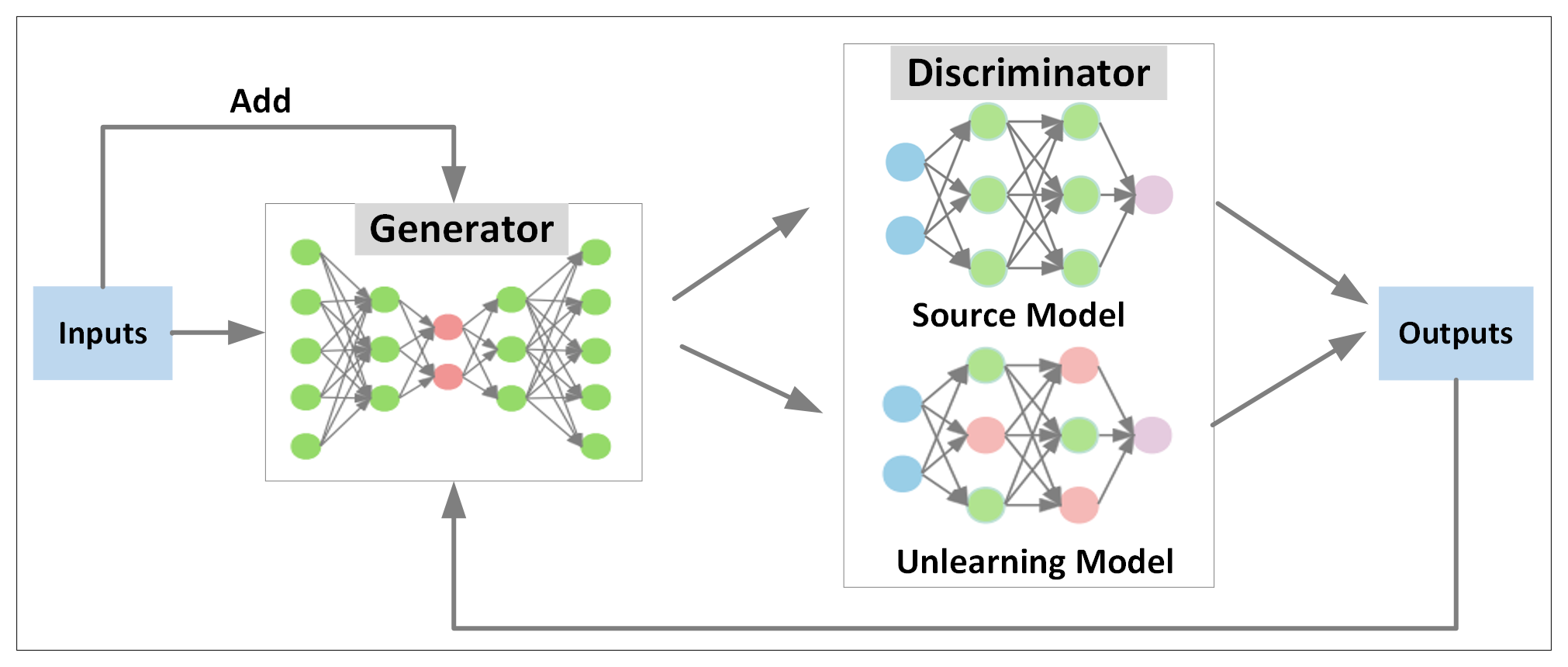}
    \caption{SPD-GAN architecture}
    \label{fig:gan}
\end{figure*}

Not all parameters affect the model equally. For instance, in image processing tasks, the earlier layers primarily focus on extracting features such as color and shape, while the later layers obtain high-dimensional information, thereby completing tasks such as object recognition. To this extent, it can be considered that perturbing the last few layers and freezing the former layers can achieve the purpose of inexact machine unlearning. 

However, only modifying the last few layers leads to a lower degree of machine unlearning, as shown in Tab.\ref{experimental results}, where the accuracy $Acc_{UL}$ on $D_{UL}$ for EU-5, EU-10, CF-5, CF-10 is significantly higher than other unlearning strategies. 

In order to make the perturbation ratio decrease and make the perturbation more precise, and to address the issue of reduced unlearning effect caused by perturbing the parameters of the last few layers. we consider identifying several parameters in the model that have the greatest impact on performance (which we call 'most sensitive') to perturb. Thus, we propose Top-K unlearning strategy.

However, the Top-K unlearning strategy usually requires sensitivity calculations, while ignoring parameters with lower sensitivity that may still contribute to the unlearning effect. Moreover, the Top-K unlearning strategy overlooks the dependencies between parameters.
Hence, we adopt a fine-grained random perturbation unlearning strategy (which we call "Random-k") ignoring these dependencies, aiming to explore the differences among the most sensitive unlearning strategy and fine-grained random perturbation strategy.

As shown in Fig.\ref{fig:framework}, the proposed unlearning strategy consists of two processes: perturbing and fine-tuning. The details include two components: the selection of perturbed parameters and the unlearning process.

\subsubsection{Parameter Selection and Perturbation}

For Random-k, by setting the perturbation ratio $k$, randomly select the model parameters to be perturbed. For Top-K, we intend to calculate the parameter sensitivity on the data set to identify the Top-K parameters that have the greatest impact on the model, so as to achieve the purpose of unlearning by performing parameter perturbation in a lighter way.

For the Top-K strategy, as previously mentioned, it is necessary to evaluate the sensitivity of each model parameter. Certainly, perturbing model parameters one by one and rank their performance change can determine the most sensitive $k$ parameters, but it's time-consuming. To quickly obtain the sensitivity of parameters, we approximate it using the gradient. A similar approach is also reflected in \cite{aljundi2018memory} for continual learning. Specifically, after the model has converged, yielding the function $F(x, w)$, we introduce a small perturbation to the model parameters denoted as $\mu = \{\mu_{i,j}\}$. The change in the model's output can be approximately measured using the following formula:

\begin{equation}
    \centering
    \bigtriangleup(x,\omega)=\sum_{i,j} \frac{\partial F(x,\omega)}{\partial \omega_{i,j}} * \mu_{i,j}
    \label{formulation4}
\end{equation}

Where $\bigtriangleup(x,w)$ represents the change in the model's output performance due to the introduced perturbation $\epsilon$, and $\partial F(x,w)/\partial\omega_{i,j} = g_{i,j}(x)$ is the gradient of learned model to parameter $w$. This evaluation of sensitivity allows us to assess the impact of each parameter on the model's predictions. 

Focusing on aforementioned Eq. \ref{formulation4}, assuming the perturbation $\epsilon$ added to the model is a same constant for all parameters, the calculation of sensitivity can be simplified as follows:

\begin{equation}
    \centering
    sensitivity_{i,j}= \| g_{i,j}(x)\|
    \label{sensitivity}
\end{equation}

By adding such a constant perturbation across all model parameters, we can directly evaluate the sensitivity of each parameter's contribution to the model's overall output change, which can be approximated by calculating the norm of the gradient. Only with one forward pass and back-propagation process, we obtain the sensitivity for each parameter. 

Fig.\ref{fig:framework} illustrates the process of our strategies. After completing the parameter screening, we introduce perturbations by drawing upon the insights from \cite{szegedy2013intriguing}. Specifically, we apply a minute amount of noise to the selected parameters as follows: 
\begin{align}
    & \omega_i = \omega_i+ \epsilon * \omega_{i} \nonumber \\
    & \omega_i \in  \{Top-K(\omega)\lor{Random-k(\omega)}\}
\end{align}

Here $\epsilon$ is a coefficient with a small value. By introducing these perturbations, the model's learned information is rendered inexact and less influential, allowing it to gradually diminish rather than being abruptly removed. This approach offers a practical and robust solution for machine unlearning, enabling models to efficiently adapt to changing data while preserving essential knowledge from previous training.

\subsubsection{Unlearning Process}

As illustrated in Fig.\ref{fig:framework}, small perturbations are introduced to the model $M$, followed by training the unlearning model $M_{UL}$ using $D_{RE}$ over multiple epochs to enable unlearning while still maintaining the performance of the remaining data. In this process, we gradually and iteratively adjust $M_{UL}$, nudging it towards a state that minimizes the influence of the unlearning data. This adaptive training strategy ensures that $M_{UL}$ retains its high predictive accuracy on the remaining data, while effectively `forgetting' the data that is removed. 

For the purpose of eliminating the influence of unlearning data on the model, retraining from scratch, despite consuming substantial computational resources and time, can achieve the optimal unlearning effect. As far as the strategy for machine unlearning is concerned, we desire that the unlearning model $M_{UL}$ approximates the retraining model $M_{RE}$, thus ensuring that the unlearning is moving in the right direction. We aim for both the retraining model and the unlearning model to display similar performance when subjected to $D_{UL}$. 

However, it clearly goes against the initial intention to perform retraining while conducting unlearning. Despite this, for models trained on i.i.d datasets, the retrained model and $M$ are quite similar, as a small amount of data is unlikely to have a profound impact on the model. Therefore, when calculating the similarity of output distributions, we use $M$ in place of the retrained model in the third step.

Jensen-Shannon (JS) divergence is used to guide the model fine-tuning of the unlearning process \cite{chundawat2023can}. As a symmetric and always finite measure, JS divergence provides a reliable metric for assessing the similarity or distance between two probability distributions. Suppose $x \in D_{UL}$, $P(\omega,x)$ and $P(\theta, x)$ are posterior distributions of retraining model $M_{RE}$ and unlearning model $M_{UL}$, respectively. JS divergence is calculated as following:

\begin{equation}
\small
    \begin{aligned}
    JS(P(\theta, x)||P(\omega,x)) = \frac{1}{2}KL\left( P(\theta, x)||\frac{P(\theta,x)+P(\omega,x)}{2} \right) \\
    + \frac{1}{2}KL\left(P(\omega, x)||\frac{P(\theta,x)+P(\omega,x)}{2}\right)
    \end{aligned}
    \label{JS}
\end{equation}

Where $KL(\cdot)$ is the Kullback-Leibler (KL) divergence. Since $P(\theta, x)$ and $P(\omega, x)$ are the output probability distributions of $M$ and $M_{UL}$ for $D_{UL}$, we have $\sum P(\theta, x) = \sum P(\omega, x) =1$, thus:

\begin{align}
   & JS(P(\theta, x)||P(\omega,x)) = -log 2+\frac{1}{2}\sum P(\theta,x)\log P(\theta,x) \nonumber \\
    & +  \frac{1}{2} \sum P(\omega,x)\log P(\omega,x) - \log(P(\theta,x) + P(\omega,x)) 
    \label{jsd}
\end{align}

In the context of machine unlearning, the JS divergence between distributions $P(\omega,x)$ and $P(\theta, x)$ can quantify the influence of $D_{UL}$ on $M_{UL}$. If the JS divergence is small, the influence of $D_{UL}$ on $M_{UL}$ is deemed to be minor. Therefore, during training, the objective is to minimize the JS divergence between these distributions to ensure the effectiveness of unlearning. Furthermore, since the JS divergence value always lies between 0 and 1, it provides an intuitive measure of the similarity between $M_{UL}$ and $M$.

Mathematically, we articulate this as an optimization challenge. A novel loss function $L$ can be postulated, encompassing the original cross entropy loss in addition to a term associated with the JS divergence:

\begin{align}
        L = & \frac{1}{|D_{RE}|} \cdot \underset{(x,y)\in D_{RE}}{\sum} \sum {y_{i,j}} \cdot log(p_{i,j}) \nonumber \\
        & + \lambda  \cdot \frac{1}{|D_{UL}|} \sum_{(x\in D_{UL})} JS(P(\theta,x)||P(\omega,x)) 
        \label{loss}  
\end{align}

Where $|D_{RE}|$ is the number of $D_{RE}$, $y_{i,j}$ is the true label of the $j-th$ class of the $i-th$ sample, and $p_{i,j}$ is the probability that the model predicts the $i-th$ sample as the $j-th$ class, $\lambda$ is used to balance the weight between the cross-entropy loss and the JS divergence.

\begin{algorithm}[h!]
	\caption{Top-K/Random-k } 
	\label{alg} 
	\begin{algorithmic}[1]
        \REQUIRE  Dataset $D$, unlearning data $D_{UL}$, remaining data $D_{RE}$, perturbed number $K$ / ratio $k$, coefficient $\epsilon$
        \STATE Train model $M$ on $D$
        \STATE $M_{UL} \gets M$    
        \IF{Top-K}
            \STATE Pick a data sample $x$ in $D$, the gradient of $\omega$: $\frac{\partial F(x,\omega)}{\partial\theta_{i,j}}$
            \FOR{$\omega_i$ in $\omega$}
                \STATE calculate sensitivity of each parameter:\\ $sensitivity_{\omega_i}\gets \| g_{\omega_i}(x)\|$
            \ENDFOR
            \STATE $\omega_j \gets \omega_j+ \epsilon \times \omega_{j}, \omega_j \in \{Top-K(\omega)\}$
        \ELSE
            \STATE $mask \gets Rand(\omega.shape)$
            \STATE $mask \gets (mask<k) \times \epsilon $
            \STATE $\omega \gets (1-mask)\times \omega + mask\times \omega_{j},j\in\{Random-k(\omega)\}$
        \ENDIF
        \WHILE{Epochs}
        \STATE $L_1\gets\frac{1}{|D_{RE}|} \cdot \sum {y_{i,j}} \cdot log(p_{i,j})$
        \STATE Computes the JS divergence of the output distributions of $M$ and $M_{UL}$ for $D_{UL}$: 
        \STATE $L_2\gets JS(P(\theta,x)||P(\omega,x)) $
        \STATE Minimize $L\gets L_1+\lambda L_2$
        \ENDWHILE
	\end{algorithmic} 
\end{algorithm}

Top-K and Random-k machine unlearning strategies are summarized in Algorithm \ref{alg}.

\subsection{Unlearning Degree Quantification} 
\label{unlearning effectiveness}
By employing a fine-grained model parameter perturbation strategy for the machine unlearning process, the unlearning model becomes nearly indistinguishable from the source model $M$. Quantifying the unlearning effect by examining the accuracy on the remaining data $D_{RE}$ and the unlearning data $D_{UL}$ of the unlearning model can be challenging. As shown in Fig.\ref{fig:iid}, the accuracy of both $D_{RE}$ and $D_{UL}$ may be maintained at a fairly satisfactory level, the difference between these two items in such experiments is about only 15\% which is quite small. Similarly, Fig.\ref{acc} confirms this observation, which is further elaborated in Section \ref{effectiveness of Top-K and Random-k}.

As mentioned earlier, we believe the reasons can be attributed to the i.i.d property of the data and the generalization ability of the model. To address the difficulty in measuring the effect of unlearning and further quantify the degree of unlearning caused by each unlearning strategy, we consider breaking the i.i.d property of $D_{UL}$.

Therefore, we design a SPD-GAN applying slight distributional perturbations to $D_{UL}$. It's necessary to ensure that the added perturbation is minimal so that the model performance difference is not too large while also ensuring that the i.i.d property is indeed broken. Therefore, the objective is that perturbed unlearning data $D_p$ is guaranteed to perform well on the source model $M$, while the performance of the perturbed unlearning data $D_p$ on the unlearning model $M_{UL}$ differs significantly from that of the remaining data $D_{RE}$ on $M_{UL}$. That is to say:
\begin{align}
    &P(M, D_p) \approx P(M, D_{UL}) \nonumber \\
    & P(M_{UL}, D_p) \ll P(M_{UL}, D_{RE})
    \label{equation9}
\end{align}

Fig.\ref{fig:gan} shows the architecture of SPD-GAN, an autoencoder serves as the generator $G(\cdot)$ for noise generation. For the discriminator, the source model $M(\cdot)$ and the unlearning model $M_{UL}(\cdot)$ are employed as a joint discriminator.

To formally represent the aforementioned process, we define the following objective function. Initially, for the discriminator, $M(\cdot)$ and $M_{UL}(\cdot)$ are already trained, they are solely used to guide the training of the generator. In order to maximize the $M(\cdot)$'s accuracy on perturbed data $D_p$ while minimizing the $M_{UL}$'s accuracy on $D_p$, for generator, we have:

\begin{equation}
\begin{split}
    L_G &= \mathbb{E}_{z \sim p_{z}(z)}[\log (1 - M(G(z)+x))] \\
    &\quad - \mathbb{E}_{z \sim p_{z}(z)}[\log M_{UL}(G(z)+x)]
\end{split}
\end{equation}

Here, $\mathbb{E}$ denotes expectation,  $z \sim p_{z}(z)$ denotes sampling from the latent variable distribution. Loss function are defined as:

\begin{equation}
    \centering
    loss = loss_{M(\cdot)} - \eta * loss_{M_{UL}(\cdot)}
    \label{spd-gan_loss}
\end{equation}

\begin{figure}
    \centering
    \includegraphics[width=0.48\textwidth]{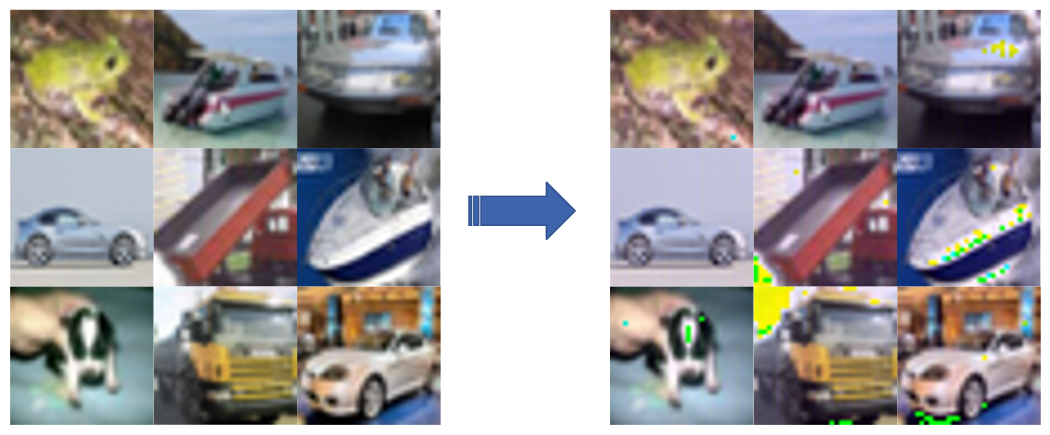}
    \caption{Visualization of images before and after SPD-GAN applied on ResNet18.}
    \label{fig:perturbed_image}
\end{figure}

Where $loss_{M(\cdot)}$ and $loss_{M_{UL}(\cdot)}$  represent the losses on $M$ and $M_{UL}$, respectively. And $\eta$ is a coefficient that balances the classification effect of $M(\cdot)$ and the unlearning effect of $M_{UL}(\cdot)$. As explained in Eq.\ref{equation1}, our goal is to explicitly demonstrate the degree of unlearning in terms of performance. Once completing the training of SPD-GAN, the performance difference between $M(\cdot)$ and $M_{UL}(\cdot)$ on perturbed data $D_p$ is indicative of unlearning degree. That is:

\begin{equation}
    \centering
    degree = P(M, D_p) - P(M_{UL}, D_p)
    \label{degree}
\end{equation}

Here, we use accuracy to represent performance $P(\cdot, \cdot)$. Specifically, we use $D_p$ as an approximation of $D_{UL}$ to evaluate the performance difference on models before and after unlearning, focusing on the non-i.i.d. property of $D_{UL}$. This difference is then defined as the degree of unlearning. Assuming total classes of data are $C$, the unlearning degree ranges in $[0, 1-1/C]$.

Unlearning degree calculating process is detailed in Algorithm \ref{spd-gan}.

\begin{algorithm}[h!]
    \caption{Unlearning Degree Evaluation} 
    \label{alg2} 
    \begin{algorithmic}[1]
        \REQUIRE Pre-trained joint discriminators $M(\cdot)$ and $M_{UL}(\cdot)$, 
        Generator $G(\cdot)$, coefficient $\eta$
        \STATE Initialize generator $G(\cdot)$ with random weights
        \WHILE{Epochs}
            \STATE Sample a batch of unlearning data $D_{UL}$ and corresponding labels $Y$
            \STATE  $D_p = G(D_{UL}) + D_{UL}$ 
            \STATE  $y_S \gets M(D_p)$ 
            \STATE  $y_U \gets M_{UL}(D_p)$
            \STATE  $loss_{M(\cdot)} \gets cross\_entropy(y_S, Y) $ 
            \STATE  $loss_{M_{UL}(\cdot)} \gets cross\_entropy(y_U, Y) $
            \STATE Minimize $loss \gets loss_{M(\cdot)}-\eta * loss_{M_{UL}(\cdot)} $
        \ENDWHILE
        \STATE $D_p=G(D_{UL})+D_{UL}$ \COMMENT{Generate final perturbed data $D_p$ for evaluation}
        \STATE $degree = P(M, D_p) - P(M_{UL}, D_p)$
    \end{algorithmic} 
    \label{spd-gan}
\end{algorithm}

\subsection{Time Complexity Analysis}
\label{complexity Analysis}

\paragraph{Theorem 1} For each epoch, the computational time of $M_{UL}$ is less than that of the retraining model $M_{RE}$.

\textit{Proof:} Let the time for a single forward pass and back-propagation of the retraining model $M_{RE}$ on the data set $D_{RE}$ be $\mathcal{O}(N)$, where $N$ denotes the model's parameter count.

Considering $M_{UL}$, its loss function can be represented as: $L(\theta) = L(\omega) + \delta L(\omega)$ with $\delta L(\omega)$ representing the change due to data removal. For the Top-K unlearning strategy, the gradient update is expressed as $\delta \omega_k= - \alpha \cdot \partial \delta L/ \partial \omega_k$ and $\delta \omega_i = 0$ for $\omega_i \notin \omega_k$, yielding a time complexity of $\mathcal{O}(K)$ for gradient recomputation. Similarly, for the Random-k strategy, the complexity is $\mathcal{O}(k\cdot N)$ with $k \in (0,1)$. Hence, strategies based on partial parameters perturbed have significantly reduced epoch time compared to full retraining since both $\mathcal{O}(K)$ and $\mathcal{O}(k\cdot N)$ is smaller than $\mathcal{O}(N)$.

\paragraph{Theorem 2} $M_{UL}$ requires fewer epochs to converge compared to the retraining model $M_{RE}$.

\textit{Proof:} Firstly, suppose $\omega'$ is the parameter of the model after perturbation, and $\omega_{rand}$ is the parameter of random initialization, thus:

\begin{equation}
    \centering
    L(\omega'; D_{RE}) \leq L(\omega_{rand}; D_{RE})
    \label{init}
\end{equation}

Where $L(\omega'; D_{RE})$ denotes the loss of the $D_{RE}$ under the parameter $\omega'$. As indicated by Eq.\ref{init}, the initial loss starting from $\omega'$ is expected to be lower.

Furthermore, when considering the magnitude and direction of the gradient $\bigtriangledown L(\omega'; D_{RE})$, since $\omega'$ represents an optimized solution, for the majority of normalized directions $d$ (where its magnitude is standardized), we have:

\begin{equation}
    \centering
    d^T \bigtriangledown L(\omega'; D_{RE}) \leq d^T \bigtriangledown L(\omega_{rand}; D_{RE})
\end{equation}

Starting from $\omega'$, the requisite step size might be reduced, implying that the gradient's direction remains more consistent.

Given that the gradient direction originating from $\omega'$ exhibits stability and that the starting loss value is already minimized, the iterative optimization is poised to converge more swiftly. Suppose the complete retraining process requires $e_1$ epochs, while Top-K machine unlearning takes $e_2$ epochs, Random-k takes $e_3$ epochs. Consequently, we deduce that: $e_1 \times \mathcal{O}(N) > e_2 \times \mathcal{O}(K)$, here both $e_1$ surpasses $e_2 $ and $\mathcal{O}(N)$ exceeds $\mathcal{O}(K)$. Same for Random-k, $e_1 \times \mathcal{O}(N) > e_3 \times \mathcal{O}(k\cdot N)$.

\paragraph{Time cost for Top-K parameters calculation}
We use Eq.\ref{sensitivity} to approximate the sensitivity of each parameter, which is the $L_2$ norm of the gradient. For each sample $x$, the time to compute the gradient and the norm is typically linear, i.e., it depends on the number of model parameters and the feature dimension of the sample. As previously noted in Theorem 1, the time for one forward pass and back-propagation is $\mathcal{O}(N)$, hence the overall time complexity for evaluating sensitivity remains $\mathcal{O}(N)$. Once the model structure is determined, the parameters with Top-K sensitivity of a specific model should be calculated only once. This means that the time required to compute the Top-K sensitive parameters will be significantly less than the time taken to perform the unlearning.

\section{Experiments}\label{experiments}
\subsection{Baseline Methods}
The impact of the first few layers on the model is mainly to extract the features of the data, and the latter few layers use these features to perform tasks. Following the setting of  \cite{DBLP:journals/corr/abs-2201-06640}, we set $K=\{5,10\}$, and use EU-K and CF-K as two of our baseline methods. To track the acceleration of machine unlearning, we also take retraining into consideration.

\textbf{EU-K}: Performing exact unlearning involves targeting the final $K$ layers of the model. In this process, the model parameters are held constant except the last $K$ layers. These last $K$ layers are then reinitialized and trained from the beginning using $D_{RE}$.

\textbf{CF-K}: Catastrophic forgetting the last $K$ layers of the model  and then fine-tuning the last $K$ layers using $D_{RE}$. Differing from EU-K, there's no need to reinitialize the parameters of the last $K$ layers. The parameters of all layers except the last $K$ are held constant.

\textbf{Retrain}: Retrain the model using $D_{RE}$ from scratch.

\subsection{Evaluation Metrics}
To adequately assess the performance of machine unlearning, we consider two key aspects: the effectiveness of the unlearning process and the model's generalization. Therefore, we propose novel metrics, namely the \emph{forgetting rate} and \emph{memory retention rate}. At the same time, we use \emph{acceleration ratio} to demonstrate the speedup and \emph{similarity} to evaluate model indistinguishability.

\paragraph{Forgetting Rate (FR)}
The forgetting rate is a measure of performance decay on the unlearning data, post the model's unlearning phase. It measures the extent of accuracy decline in the unlearning data after machine unlearning, relative to the accuracy before unlearning.

  \begin{equation}
      \begin{aligned}
          FR = \frac{Acc_{before}-Acc_{after}}{Acc_{before}}
      \end{aligned}
  \end{equation}

Here, $Acc_{before}$ refers to the accuracy of model $M$ on the unlearning data before the unlearning process is applied, while $Acc_{after}$ denotes the accuracy of model $M_{UL}$ on the same data set after the unlearning process has been completed.

\paragraph{Memory Retention Rate (MRR)}
In machine unlearning, it's also crucial to assess the performance of the unlearning model on the remaining data. To this end, we employ the memory retention rate:

\begin{equation}
    \begin{aligned}
        MRR =  \frac{Acc'_{after}}{Acc_{before}}
    \end{aligned}
\end{equation}

Here $Acc'_{after}$ represents the accuracy of the unlearning model on the remaining data. 

\paragraph{Acceleration ratio} In the preceding text, we analyzed the time complexity of the machine unlearning strategy based on partial parameter perturbation. We utilize `unlearn time' to measure the time taken for the unlearning model to reach convergence and calculate the acceleration ratio relative to the retraining method.

\paragraph{Similarity} To assess the degree of model indistinguishability, we employ the $similarity = 1-JS~divergence$ as a metric to quantify the similarity between the unlearning model and the retraining model, where JS divergence is defined in Eq.\ref{jsd}.

\subsection{Experimental Results}
\subsubsection{Training settings}

We conduct experiments on CIFAR-10 \cite{krizhevsky2009learning} using ResNet18 \cite{he2016deep}, VGG \cite{simonyan2014very}, GoogLeNet \cite{szegedy2015going}, and DenseNet \cite{huang2017densely} models. The training dataset is randomly divided into $D_{UL}$ and $D_{RE}$ with $D_{UL}$ comprising 5\%, 10\%, 15\% and 20\%. For Top-K, we set $K=45$ for ResNet18, and for the Random-k, $k=5\%$. The coefficient $\lambda$ in Eq.\ref{loss} is set to 0.1. We use Adam as optimizer, for EU-K and CF-K, we set learning rate $1e-4$. The generator architecture in SPD-GAN consists of (two Conv2d)-BatchNorm-Conv2d-BatchNorm layers for the encoder, followed by ConvTranspose2d-BatchNorm-ConvTranspose2d-Conv2d layers for the decoder. The LeakyReLU and ReLU activation functions are used for the encoder and decoder, respectively. Activation function Tanh is used to limit the output within [-1, 1]. The coefficient $\eta$ in Eq.\ref{spd-gan_loss} is 0.03. All the experiments are conducted on a Nvidia A100 GPU.

\subsubsection{Determination of $K$ value for Top-K}

\begin{figure}[h!]
    \centering
\subfigure[Unlearning data ratio of 5\%]{
    \begin{minipage}[t]{0.46\linewidth}
    \label{k_5}
    \centering
    \includegraphics[width=1\textwidth]{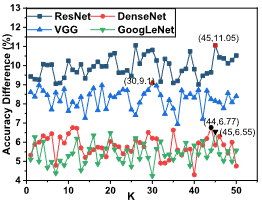}
    \end{minipage}
}%
\subfigure[Unlearning data ratio of 10\%]{
    \begin{minipage}[t]{0.46\linewidth}
    \label{k_10}
    \centering
    \includegraphics[width=1\textwidth]{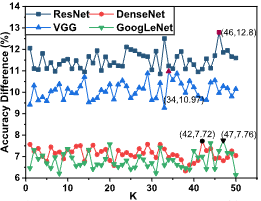}
    \end{minipage}
}

\subfigure[Unlearning data ratio of 15\%]{
    \begin{minipage}[t]{0.46\linewidth}
    \centering
    \label{k_15}
    \includegraphics[width=1\textwidth]{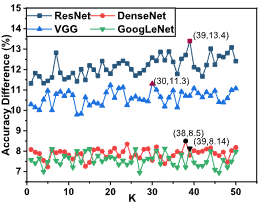}
    \end{minipage} 
}
\subfigure[Unlearning data ratio of 20\%]{
    \begin{minipage}[t]{0.46\linewidth}
    \label{k_20}
    \centering
    \includegraphics[width=1\textwidth]{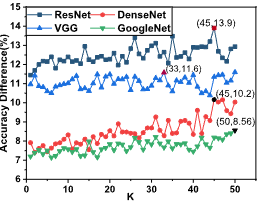}
    \end{minipage}
}
\caption{Accuracy difference between $D_{RE}$ and $D_{UL}$ for Top-K under unlearning data ratio at (a) 5\%, (b) 10\%, (c) 15\% and (d) 20\%. The maximum accuracy difference and its corresponding K value is annotated. Such K value is identified as the optimal. }
\label{acc_topk}
\end{figure}

\begin{figure}[h!]
  \centering
  \subfigure[ResNet18]{
    \begin{minipage}[t]{0.46\linewidth}
    \label{ResNet_epochs}
    \centering
    \includegraphics[width=1\textwidth]{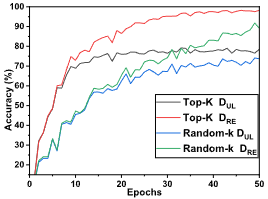}
    \end{minipage}
    }
    \subfigure[VGG]{
    \begin{minipage}[t]{0.46\linewidth}
    \label{vgg_epochs}
    \centering
    \includegraphics[width=1\textwidth]{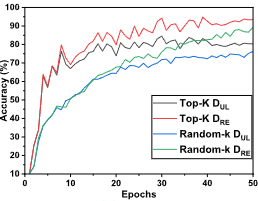}
    \end{minipage}
    } 
    
    \subfigure[DenseNet]{
    \begin{minipage}[t]{0.46\linewidth}
    \label{DenseNet_epochs}
    \centering
    \includegraphics[width=1\textwidth]{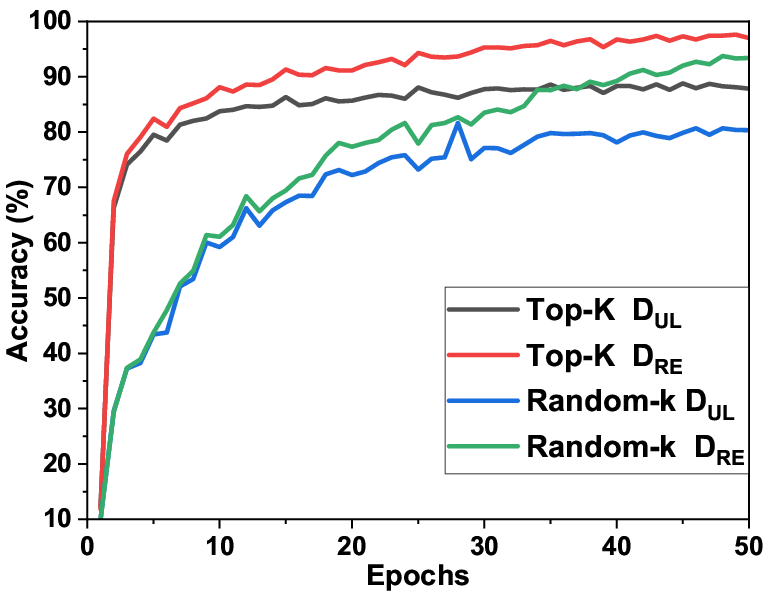}
    \end{minipage}
    }
    \subfigure[GoogLeNet]{
    \begin{minipage}[t]{0.46\linewidth}
    \label{GoogLeNet_epochs}
    \centering
    \includegraphics[width=1\textwidth]{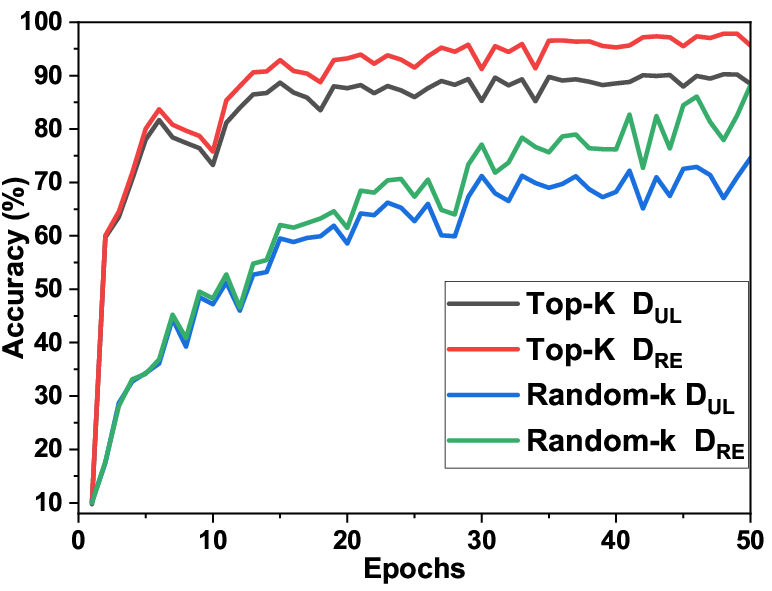}
    \end{minipage}
    }
  \caption{Top-K/Random-k unlearning accuracy (\%) on (a) ResNet18, (b) VGG, (c) DenseNet and (d) GoogLeNet.}
  \label{acc}
\end{figure}

\begin{figure*}[h!]
    \centering
    \subfigure[ResNet18]{
    \begin{minipage}[t]{0.23\linewidth}
    \centering
    \includegraphics[width=1\textwidth]{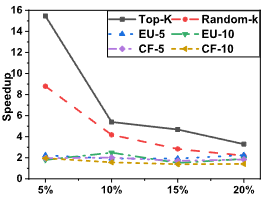}
    \end{minipage}
    \label{speed_resnet}
}%
\subfigure[VGG]{
    \begin{minipage}[t]{0.23\linewidth}
    \centering
    \includegraphics[width=1\textwidth]{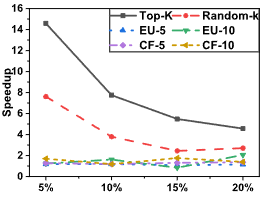}
    \end{minipage}
    \label{speed_vgg}
}%
\subfigure[DenseNet]{
    \begin{minipage}[t]{0.23\linewidth}
    \centering
    \includegraphics[width=1\textwidth]{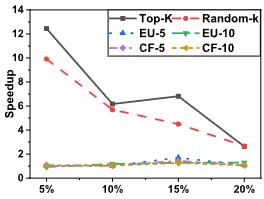}
    \end{minipage}
    \label{speed_densenet}
}
\subfigure[GoogLeNet]{
    \begin{minipage}[t]{0.23\linewidth}
    \centering
    \includegraphics[width=1\textwidth]{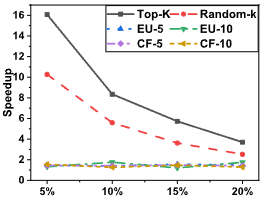}
    \end{minipage}
    \label{speed_googlenet}
}
\caption{Acceleration compared to retraining under different unlearning ratio on (a) ResNet18, (b) VGG, (c) DenseNet and (d) GoogLeNet}
\label{fig:speedup}
\end{figure*}

\begin{table}[!h]
\centering
\caption{The number of parameters perturbed in different models under various unlearning strategies.}
\begin{tabular}{c|cccc}
\hline
Strategy    & ResNet18 & VGG & DenseNet & GoogLeNet \\ \hline
Top-K       & \textbf{45}       & \textbf{30}  & \textbf{45}       & \textbf{50}        \\ \hline
Random-k    & 0.558M   &  1.002M   & 0.348M         &  0.308M        \\ \hline
EU-5/CF-5   & 2.37M    &  6666   &   49162       & 10634          \\ \hline
EU-10/CF-10 & 4.73M    & 4.73M    & 0.178M         &  0.265M         \\ \hline
\end{tabular}

\label{number}
\end{table}

\begin{table*}[h!]
\centering
\caption{The experimental results of the ResNet18 under different unlearning ratios of the CIFAR-10. Lower $Acc_{UL} (\%)$ and Time (s) and higher other metrics (\%) indicate better performance.}
\resizebox{0.85\textwidth}{!}{%
\begin{tabular}{c|ccccccc}
\hline
$|D_{UL}|/|D|$ & Strategy & $Acc_{UL}$ & $Acc_{RE}$ & FR & MRR & Similarity & Unlearn Time \\ \hline
\multirow{7}{*}{5\%} & Top-K & 80.01\% & 97.61\% & 19.97\% & 97.61\% & 88.37\% & \textbf{76.97s} \\
& Random-k & \textbf{76.56\%} & 90.07\% & \textbf{23.44\%} & 90.07\% & 78.77\% & 135.43s \\
& EU-5 & 90.68\% & 98.44\% & 9.30\% & 98.44\% & 84.88\% & 534.76s \\
& EU-10 & 88.88\% & 99.02\% & 11.11\% & 99.02\% & 69.42\% & 662.34s \\
& CF-5 & 86.52\% & 99.16\% & 13.46\% & 99.18\% & 78.65\% & 586.6s \\
& CF-10 & 86.64\% & 99.54\% & 13.34\% & 99.56\% & 99.98\% & 607.24s \\
& Retrain & 84.84\% & 100\% & 15.16\% & 100\% & 100\% & 1190.15s \\ \hline
\multirow{7}{*}{10\%} & Top-K & \textbf{80.14\%} & 97.19\% & \textbf{19.84\%} & 97.21\% & 85.36\% & \textbf{192.86s} \\
& Random-k & 81.48\% & 96.52\% & 18.52\% & 96.54\% & 91.03\% & 249.11s \\
& EU-5 & 88.7\% & 98.88\% & 11.28\% & 98.88\% & 82.25\% & 541.38s \\
& EU-10 & 86.88\% & 98.12\% & 13.12\% & 98.14\% & 90.4\% & 417.87s \\
& CF-5 & 85.52\% & 97.89\% & 14.46\% & 97.91\% & 70.27\% & 515.13s \\
& CF-10 & 86.26\% & 98.91\% & 13.73\% & 98.93\% & 91.37\% & 664.39s \\
& Retrain & 84.8\% & 100\% & 15.2\% & 100\% & 100\% & 1040.08s \\ \hline
\multirow{7}{*}{15\%} & Top-K & \textbf{75.48\%} & 96.8\% & \textbf{24.52\%} & 96.82\% & 89.88\% & \textbf{203.23s} \\
& Random-k & 79.85\% & 98.09\% & 20.13\% & 98.11\% & 83.59\% & 335.87s \\
& EU-5 & 84.61\% & 97.46\% & 15.37\% & 97.48\% & 85.11\% & 494.60s\\
& EU-10 & 83.52\% & 98.75\% & 16.46\% & 98.77\% & 91.38\% & 627.94s \\
& CF-5 & 83.64\% & 95.46\% & 16.34\% & 95.48\% & 85.27\% & 558.31s \\
& CF-10 & 82.05\% & 94.79\% & 17.93\% & 94.79\% & 91.75\% & 685.18s \\
& Retrain & 84.85\% & 100\% & 15.15\% & 100\% & 100\% & 950.51s \\ \hline
\multirow{7}{*}{20\%} & Top-K & \textbf{75.46\%} & 97.07\% & \textbf{24.54\%} & 97.07\% & 86.37\% & \textbf{425.01s} \\
& Random-k & 77.83\% & 98.99\% & 22.17\% & 98.89\% & \textbf{94.51\%} & 638.63s \\
& EU-5 & 84.48\% & 97.91\% & 15.51\% & 97.91\% & 71.77\% & 636.77s \\
& EU-10 & 83.19\% & 96.81\% & 16.79\% & 96.81\% & 83.92\% & 741.51s \\
& CF-5 & 82.86\% & 93.14\% & 17.12\% & 93.14\% & 81.76\% & 778.28s \\
& CF-10 & 81.69\% & 92.17\% & 18.29\% & 92.12\% & 91.9\% & 979.65s \\
& Retrain & 85.24\% & 100\% & 14.76\% & 100\% & 100\% & 1400.9s \\ \hline
\end{tabular}
}
\label{experimental results}
\end{table*}

We aim to achieve the effect of inexact machine unlearning by minimal number of parameter perturbed, to investigate the optimal value of \( K \) for Top-K unlearning strategy, we conduct experiments with different values of \( K \) within the range of 0 to 50 when the unlearning data ratio varies from 5\% to 20\%. Following the above settings, we apply Top-K to four models over 100 epochs. A metric to determine the optimal value of $K$ is necessary. Since machine unlearning aims to eliminate the influence of unlearning data while trying to not hurt the performance of remaining data, we calculate the accuracy difference between $Acc_{RE}$ and $Acc_{UL}$. As illustrated in Fig.\ref{acc_topk}, the \( K \) value corresponding to the maximum accuracy difference is identified as the optimal. As shown in Tab.\ref{number}, considering the optimal $K$ values under four unlearning data ratios, we set \( K=45 \) for ResNet18 and DenseNet, \( K=30 \) for VGG, and \( K=50 \) for GoogLeNet. 

As shown in Fig.\ref{acc_topk}, although we have chosen the $K$ corresponding to the maximum accuracy difference, in actual experiments, the selection of $K$ values fluctuating within the range of 0 to 50 does not bring significant variability to the Top-K unlearning. For example, in Fig.\ref{k_5}, for ResNet18, setting K=25 versus 45 does not make a substantial difference. Additionally, measuring the difference in accuracy is a sufficient but not necessary condition for the effectiveness of machine unlearning. In other words, for a fixed model, dynamically adjusting the value of $K$ based on the unlearning data ratio is highly impractical. Therefore, additional efforts or computational costs to determine a very precise $K$ value are unnecessary; we can roughly set it to a value such as 45.

\subsubsection{Effectiveness of Top-K and Random-k \label{effectiveness of Top-K and Random-k}}
Fig.\ref{acc} illustrates the unlearning accuracy $Acc_{UL}$ and remaining accuracy $Acc_{RE}$ of four different models trained on the CIFAR-10 using Top-K and Random-k over 50 epochs with unlearning ratio 20\%. At the beginning of fine-tuning, in terms of accuracy, the $D_{RE}$ corresponding to Top-K is significantly higher than that of Random-k, indicating that fewer parameter perturbations result in a better memory retention effect. Conversely, the $D_{UL}$ corresponding to Random-k is significantly lower than that of Top-K, suggesting that more parameter perturbations can enhance the extent to which the model's classification ability on $D_{UL}$ is disrupted. 

Fig.\ref{fig:iid} further shows the accuracy after applying the Random-k on the CIFAR-10 for 50 epochs. The accuracy of $D_{RE}$ is significantly higher than that of $D_{UL}$. As shown in Fig.\ref{acc}, it can be observed that the application of the Top-K achieves  accuracy close to 95\% ($Acc_{RE}$) across the four models, indicating a strong memory retention capability. Meanwhile, the accuracy difference between unlearning model on the $D_{RE}$ and the $D_{UL}$ (both in Fig.\ref{acc} and Fig.\ref{fig:iid}) shows that both the Top-K and Random-k achieve the unlearning effect more or less.

\begin{figure*}[h!]
    \centering
    \subfigure[5\% on ResNet18]{
    \begin{minipage}[t]{0.24\linewidth}
    \label{RT_resnet5}
    \centering
    \includegraphics[width=1\textwidth]{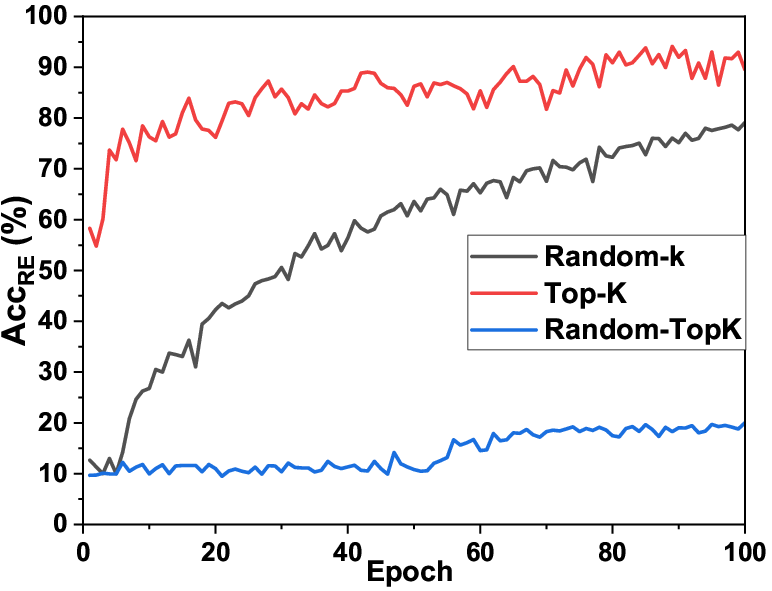}
    \end{minipage}
}%
\subfigure[10\% on ResNet18]{
    \begin{minipage}[t]{0.24\linewidth}
    \label{RT_resnet10}
    \centering
    \includegraphics[width=1\textwidth]{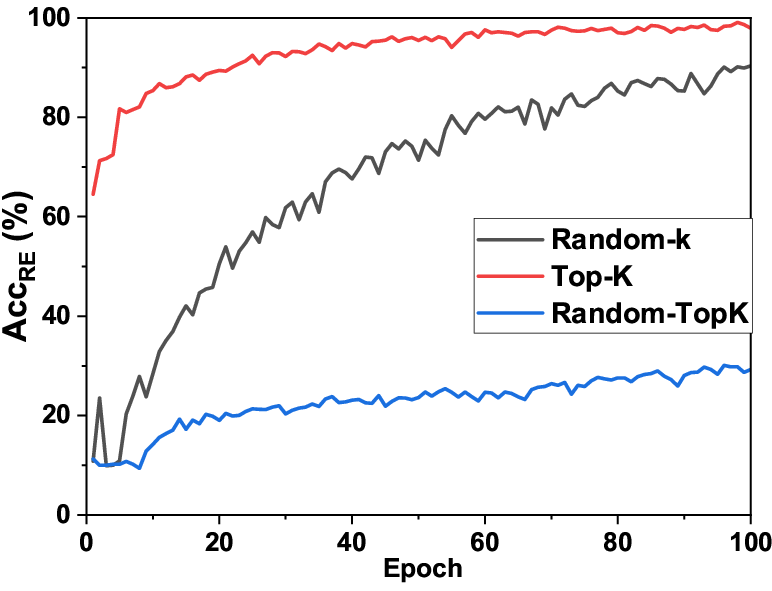}
    \end{minipage}
}%
\subfigure[15\% on ResNet18]{
    \begin{minipage}[t]{0.24\linewidth}
    \label{RT_resnet15}
    \centering
    \includegraphics[width=1\textwidth]{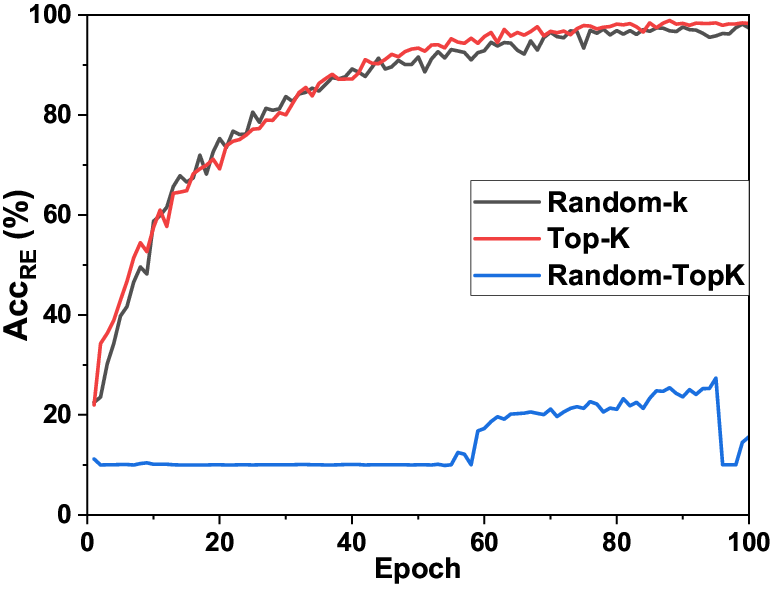}
    \end{minipage}
}%
\subfigure[20\% on ResNet18]{
    \begin{minipage}[t]{0.24\linewidth}
    \label{RT_resnet20}
    \centering
    \includegraphics[width=1\textwidth]{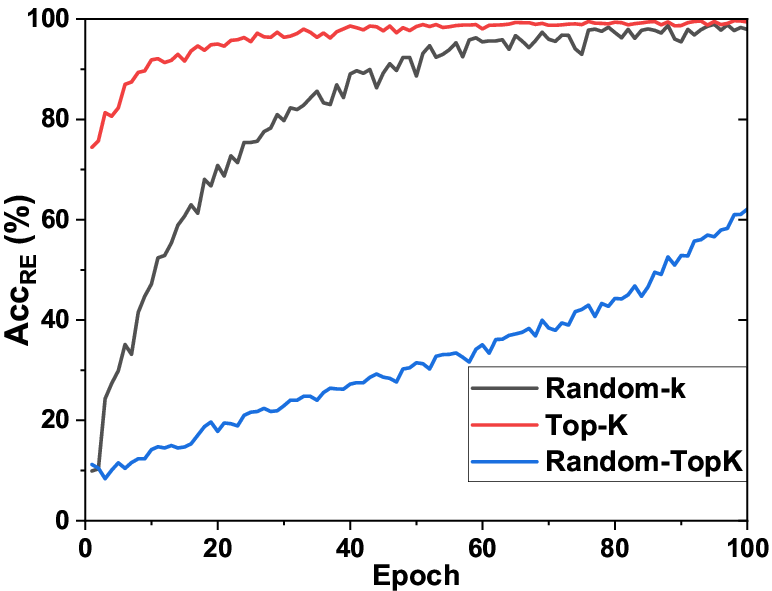}
    \end{minipage}
}%

    \subfigure[5\% on VGG]{
    \begin{minipage}[t]{0.24\linewidth}
    \label{RT_vgg5}
    \centering
    \includegraphics[width=1\textwidth]{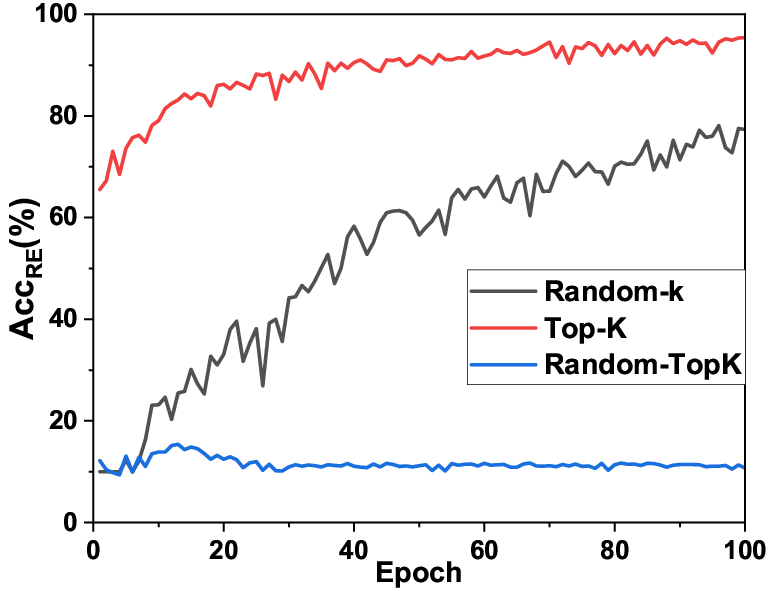}
    \end{minipage}
}%
\subfigure[10\% on VGG]{
    \begin{minipage}[t]{0.24\linewidth}
    \label{RT_vgg10}
    \centering
    \includegraphics[width=1\textwidth]{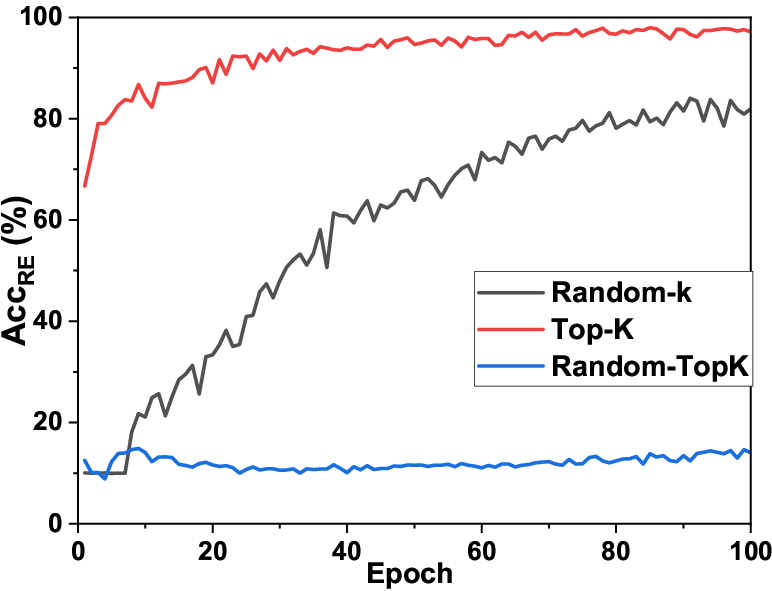}
    \end{minipage}
}%
\subfigure[15\% on VGG]{
    \begin{minipage}[t]{0.24\linewidth}
    \label{RT_vgg15}
    \centering
    \includegraphics[width=1\textwidth]{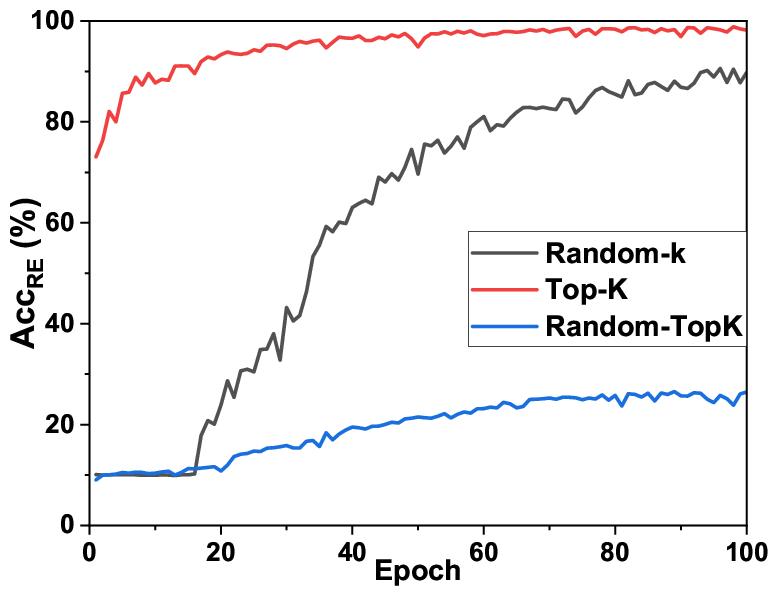}
    \end{minipage}
}%
\subfigure[20\% on VGG]{
    \begin{minipage}[t]{0.24\linewidth}
    \label{RT_vgg20}
    \centering
    \includegraphics[width=1\textwidth]{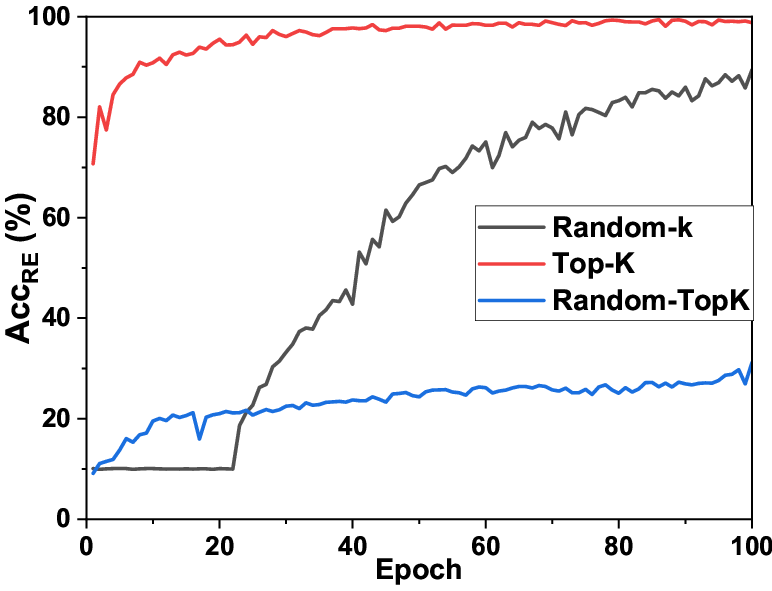}
    \end{minipage}
}%

    \subfigure[5\% on DenseNet]{
    \begin{minipage}[t]{0.24\linewidth}
    \label{RT_densenet5}
    \centering
    \includegraphics[width=1\textwidth]{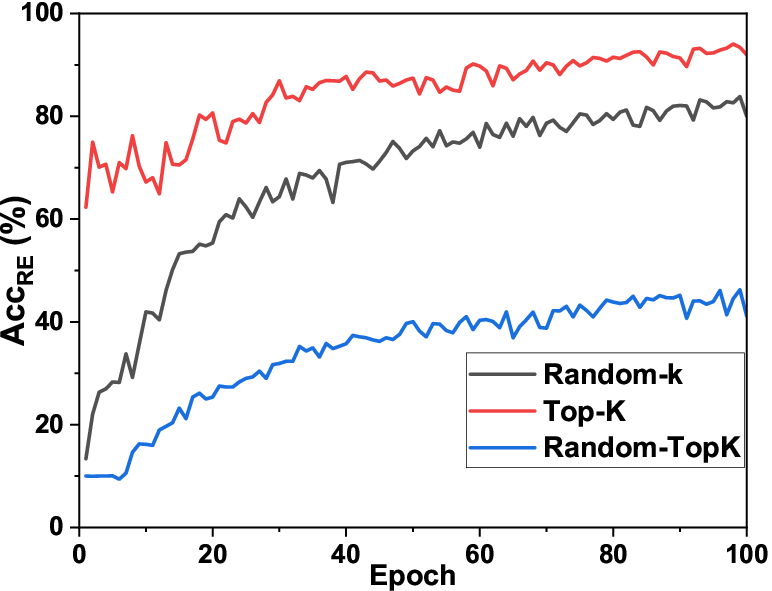}
    \end{minipage}
}%
\subfigure[10\% on DenseNet]{
    \begin{minipage}[t]{0.24\linewidth}
    \label{RT_densenet10}
    \centering
    \includegraphics[width=1\textwidth]{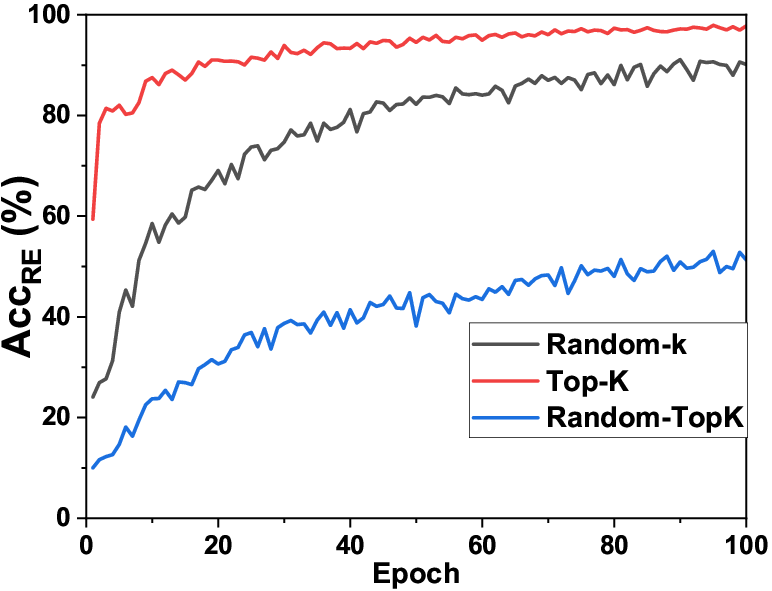}
    \end{minipage}
}%
\subfigure[15\% on DenseNet]{
    \begin{minipage}[t]{0.24\linewidth}
    \label{RT_densenet15}
    \centering
    \includegraphics[width=1\textwidth]{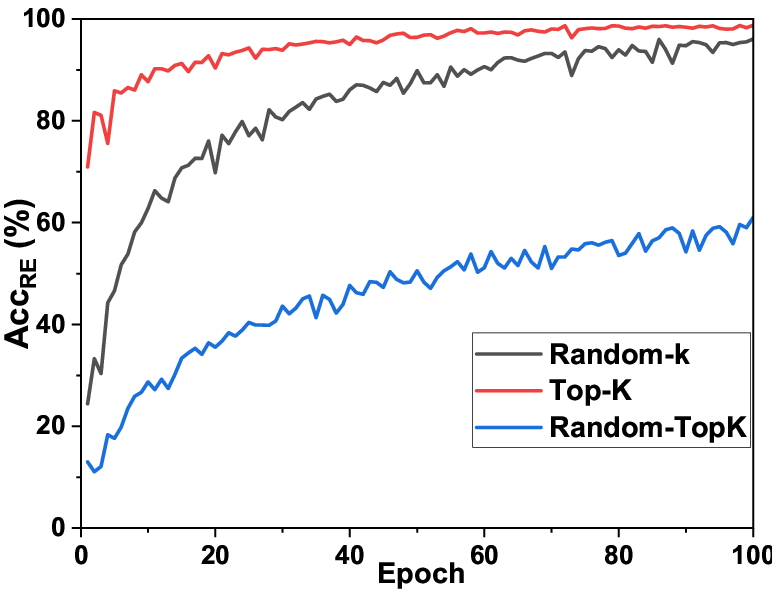}
    \end{minipage}
}%
\subfigure[20\% on DenseNet]{
    \begin{minipage}[t]{0.24\linewidth}
    \label{RT_densenet20}
    \centering
    \includegraphics[width=1\textwidth]{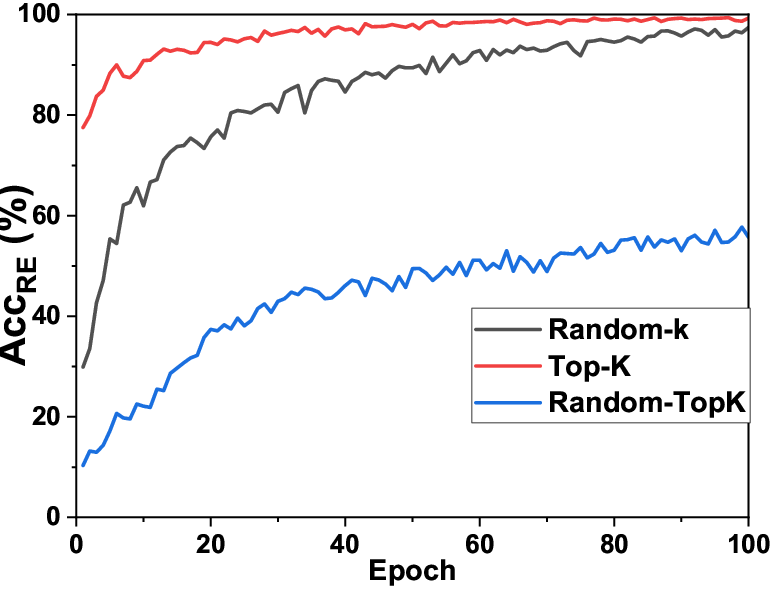}
    \end{minipage}
}%

    \subfigure[5\% on GoogLeNet]{
    \begin{minipage}[t]{0.24\linewidth}
    \label{RT_googlenet5}
    \centering
    \includegraphics[width=1\textwidth]{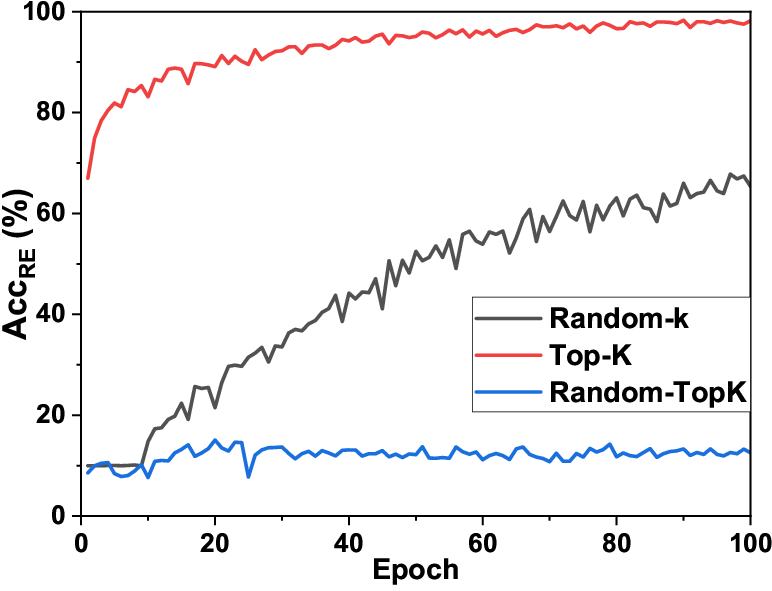}
    \end{minipage}
}%
\subfigure[10\% on GoogLeNet]{
    \begin{minipage}[t]{0.24\linewidth}
    \label{RT_googlenet10}
    \centering
    \includegraphics[width=1\textwidth]{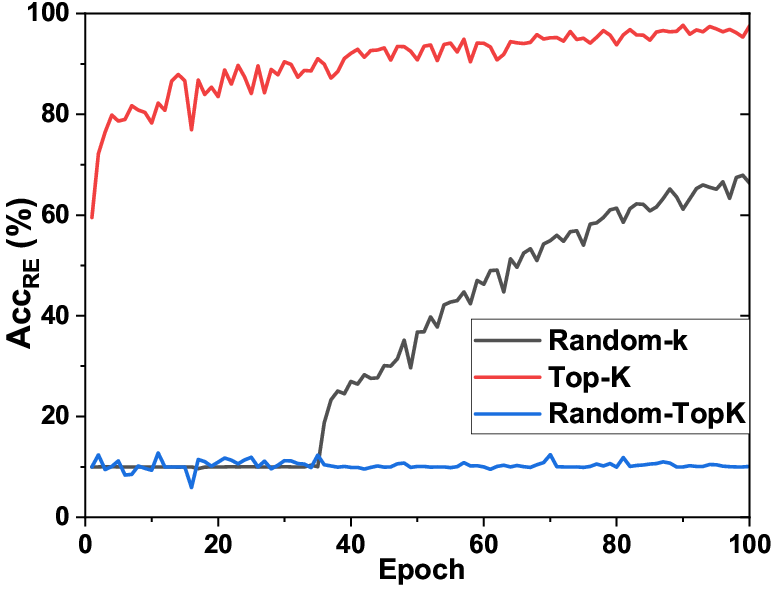}
    \end{minipage}
}%
\subfigure[15\% on GoogLeNet]{
    \begin{minipage}[t]{0.24\linewidth}
    \label{RT_googlenet15}
    \centering
    \includegraphics[width=1\textwidth]{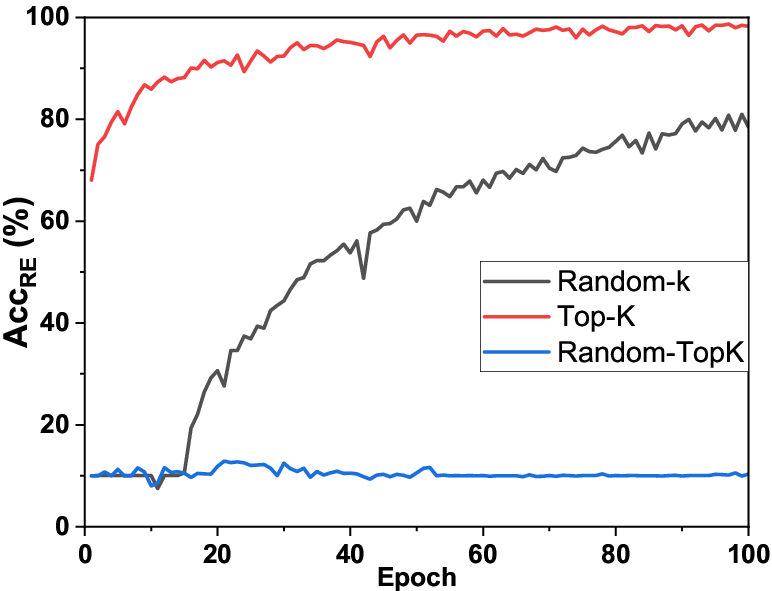}
    \end{minipage}
}%
\subfigure[20\% on GoogLeNet]{
    \begin{minipage}[t]{0.24\linewidth}
    \label{RT_googlenet20}
    \centering
    \includegraphics[width=1\textwidth]{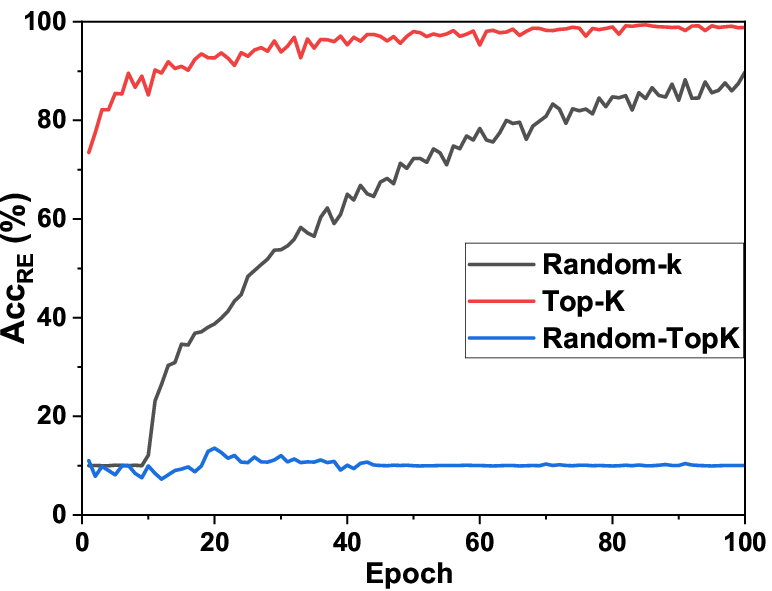}
    \end{minipage}
}%

\caption{ $Acc_{RE}$ (\%) comparison using Random-k, Top-K and mixed Random-TopK under different unlearning data ratios in [5\%, 10\%, 15\%, 20\%] on (a-d) ResNet18, (e-h) VGG, (i-l) DenseNet and (m-p) GoogLeNet.}
\label{RT}
\end{figure*}

Tab.\ref{experimental results} provides a more comprehensive overview of the effects of different unlearning strategies at various unlearning ratios ($|D_{UL}|/|D|$). Across different unlearning ratios, Top-K consistently exhibits the lowest $Acc_{UL}$ values (80.14\%, 75.48\%, and 75.46\%), along with corresponding forgetting rates (19.84\%, 24.52\%, and 24.54\%). This suggests that, relative to other strategies, Top-K achieves the most effective reduction of the impact of $D_{UL}$ on the model when the unlearning ratio belongs to 10\%, 15\%, 20\%. However, for the 5\% unlearning ratio under Random-k, its corresponding $Acc_{RE}$ and MRR of 90.07\% noticeably lag behind the results of other strategies. We believe this disparity is due to the instability of the strategy that employs random parameter perturbation, leading to inconsistent unlearning effects. In the meanwhile, across both $Acc_{RE}$ and MRR, the performance of $M_{UL}$ from nearly all unlearning strategies is comparable to that of retraining models, achieving close to 100\% accuracy. As for similarity, there isn't a significant distinction among different strategies, but this does indicate that model indistinguishability is maintained across diverse strategies.

As the ratio of $D_{UL}$ increases, the model becomes more attuned to the characteristics of the $D_{UL}$, leading to a further reduction in unlearning accuracy ($Acc_{UL}$) under an ideal unlearning strategy. However, as the ratio of $D_{RE}$ decreases, the model's discriminative ability for $D_{RE}$ ($Acc_{RE}$) may also decline. As shown in Tab.\ref{experimental results}, across varying unlearning ratios, Top-K and Random-k exhibit comparable $Acc_{RE}$ within their respective settings, while as the unlearning ratio increases, the unlearning accuracy $Acc_{UL}$ continues to decrease (from 80.01\% to 75.46\% for Top-K).

For EU-K, fine-tuning the last $K$ layers to eliminate the influence of $D_{UL}$ allows the model to leverage the unaffected features from the frozen earlier layers for better predictions. At higher unlearning ratios, this strategy results in a decrease in $Acc_{UL}$ (from 90.68\% to 84.48\% for EU-5), while the performance on $D_{RE}$ remains stable.

\begin{figure*}[h!]
    \centering
    \subfigure[5\%]{
    \begin{minipage}[t]{0.23\linewidth}
    \centering
    \includegraphics[width=1\textwidth]{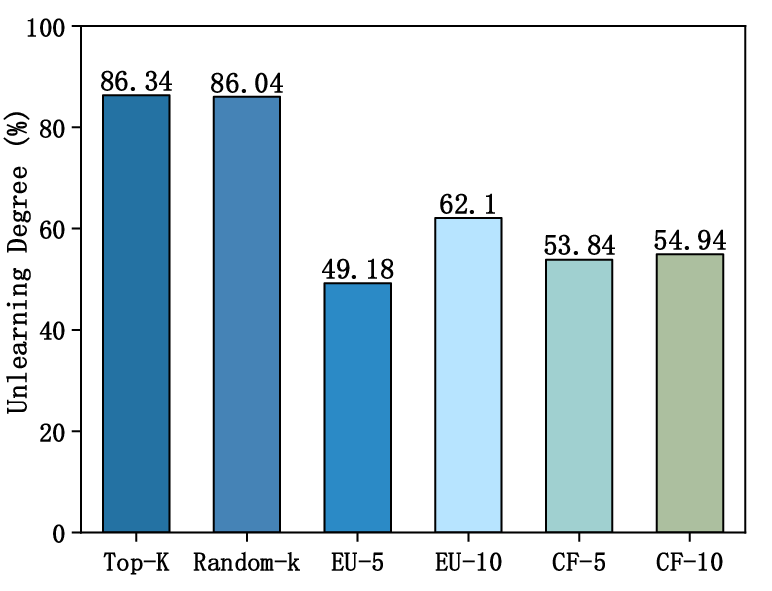}
    \end{minipage}
    \label{degree_5}
}%
\subfigure[10\%]{
    \begin{minipage}[t]{0.23\linewidth}
    \centering
    \includegraphics[width=1\textwidth]{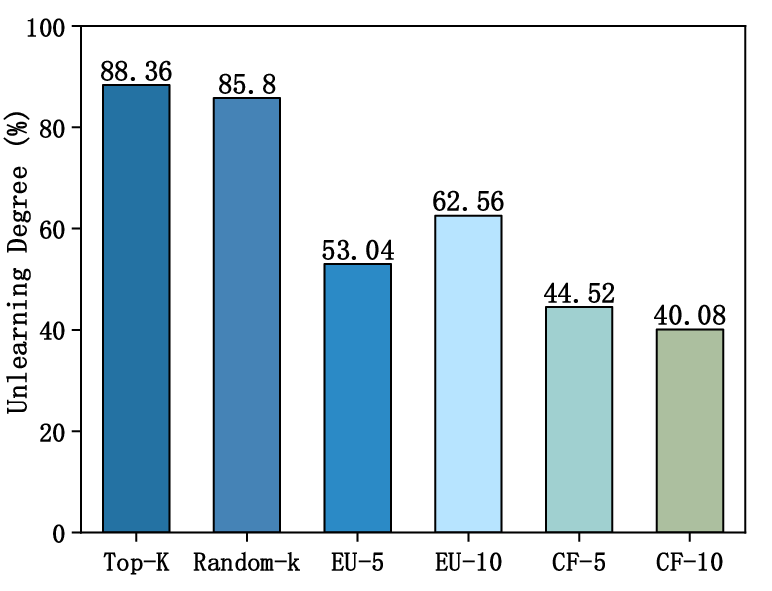}
    \end{minipage}
    \label{degree_10}
}%
\subfigure[15\%]{
    \begin{minipage}[t]{0.23\linewidth}
    \centering
    \includegraphics[width=1\textwidth]{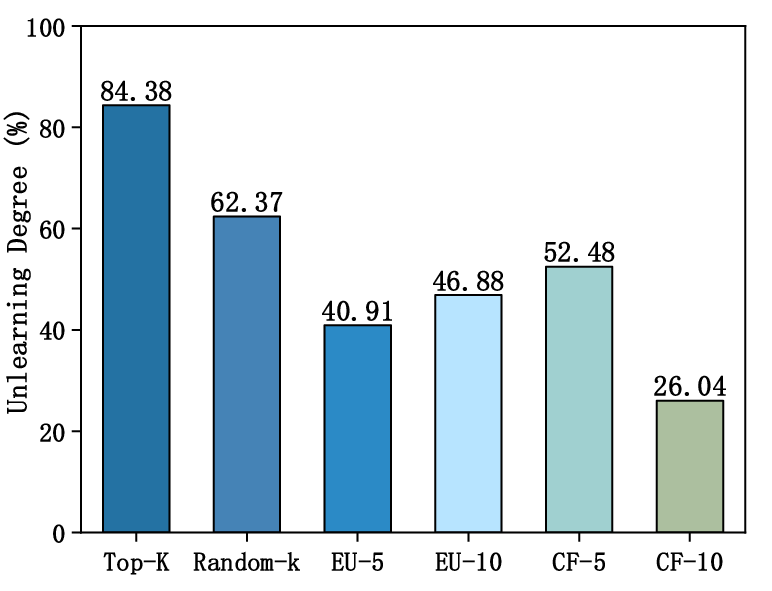}
    \end{minipage}
    \label{degree_15}
}
\subfigure[20\%]{
    \begin{minipage}[t]{0.23\linewidth}
    \centering
    \includegraphics[width=1\textwidth]{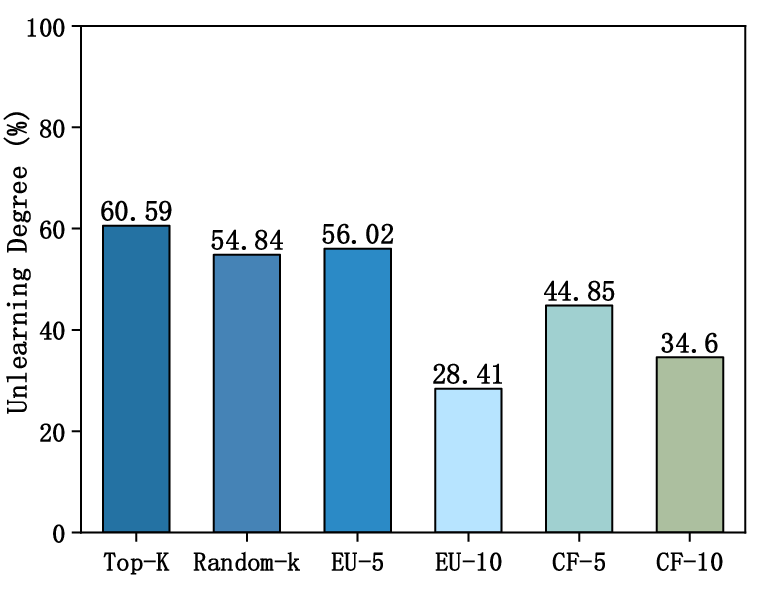}
    \end{minipage}
    \label{degree_20}
}
\caption{Unlearning degree (\%, ranges in [0, 90\%] for CIFAR-10) evaluation for different unlearning strategies on ResNet18 using CIFAR-10 under different unlearning data ratio ((a)5\%, (b)10\%, (c)15\% and (d)20\%). The higher unlearning degree is regarded as the better.}
\label{unlearning degree}
\end{figure*}

In contrast, for CF-K, the parameters of the last $K$ layers are reinitialized, granting the model enhanced flexibility to relearn and mitigate the influence of unlearning data. This leads to a continuous decline in $Acc_{UL}$ (from 86.52\% to 82.86\% for CF-5). However, reinitializing and retraining the last $K$ layers negatively impacts the model's performance on the $D_{RE}$. As reflected in Tab.\ref{experimental results}, under the CF-K strategy, $Acc_{RE}$ also shows a declining trend as the unlearning ratio increases (from 99.16\% to 93.14\% for CF-5).

As discussed earlier, we believe that employing machine unlearning strategies offers a faster way to eliminate the influence of unlearning data on models compared to the retraining from scratch. The metric 'Unlearn Time' in Tab.\ref{experimental results} indicates the time required for various strategies to complete unlearning. All strategies achieve an accelerations, but Top-K and Random-k take less time. As illustrated in Tab.\ref{number}, this acceleration is due to the fact that, for experiments on ResNet18, Top-K and Random-k perturb less parameters than others. 

To further illustrate the acceleration effect of unlearning strategies on different models compared to the retraining method, we also conduct machine unlearning tasks on VGG, DenseNet, and GoogLeNet. Fig.\ref{fig:speedup} shows the acceleration ratio of each unlearning strategy compared to retraining under different models and unlearning ratios. The acceleration effect of Top-K and Random-k is better than other strategies. Taking Fig.\ref{speed_resnet} as an example, when the unlearning ratio is 5\%, Top-K can achieve an acceleration of more than 15$\times$, while Random-k achieves an acceleration close to 9$\times$. The acceleration from EU-K and CF-K methods is not significant. As the unlearning ratio increases, the acceleration of Top-K and Random-k slows down. This is because, as more data is to be forgotten, the degree of modification to the model deepens. In conclusion, we can ascertain that the Top-K achieves optimal unlearning performance with minimal parameter perturbed, maintains considerable memory retention capability, and achieves the fastest acceleration effect.

\subsubsection{Mixed Top-K and Random-k}

In Random-k, parameters are chosen randomly to be perturbed, while in Top-K, parameters to be perturbed are carefully calculated. When the value of k is quite small, it is likely that the unlearning process only modifies redundant neuron weights, failing to achieve effective unlearning. To investigate whether the random selection in Random-k is efficient enough or can be further improved, we designed a set of experiments in which $K$ in $k\%$ parameters selected by Random-k is replaced by those chosen by Top-K. This formed a mixed Random-TopK unlearning strategy, and related experiments are conducted using settings similar to the aforementioned ones. The $Acc_{RE}$ for Top-K, Random-k, and Random-TopK after 100 training epochs are shown in Fig.\ref{RT}.

Based on the $Acc_{RE}$ trends for Top-K, Random-k, and Random-TopK, at the beginning of training, the $Acc_{RE}$ for Top-K is higher than that for Random-k and Random-TopK. This implies that as the number of perturbed parameters in the model increases, more knowledge is eliminated. For Random-TopK, its $Acc_{RE}$ is lower than both Top-K and Random-k, indicating that over-modifying parameter based on those with the highest sensitivity will be more time-consuming. For models like ResNet18 (Fig.\ref{RT_resnet5}-\ref{RT_resnet20}), VGG (Fig.\ref{RT_vgg5}-\ref{RT_vgg20}), and DenseNet (Fig.\ref{RT_densenet5}-\ref{RT_densenet20}), Random-TopK still exhibits a certain upward trend in $Acc_{RE}$, suggesting that even with excessive parameter modifications, Random-TopK can still achieve partial unlearning effects. However, we notice that when using the Random-Topk stragegy on GoogLeNet (Fig.\ref{RT_googlenet5}-\ref{RT_googlenet20}), the model fell into a state of complete non-learning. We believe this may be due to the different sensitivities and parameter dependencies between the different branches of the Inception module in GoogLeNet. Employing the Random-TopK strategy might introduce complexities, adversely affecting certain branches.

\subsubsection{Unlearning Degree Evaluation \label{unlearning degree evaluation}}
Fig.\ref{fig:perturbed_image} shows the original images and the perturbed images when SPD-GAN is applied on ResNet18. When noise is added to $D_{UL}$, it doesn't impact people's ability to correctly recognize images. However, it does disrupt the $D_{UL}$ distribution.  

As aforementioned, in order to measure the unlearning degree for each unlearning strategy, we use Eq.\ref{degree} to calculate the performance difference on the perturbed data $D_p$ generated by SPD-GAN between the source model $M$ and unlearning model $M_{UL}$. Utilizing source model $M$ and unlearning model $M_{UL}$ obtained from applying different unlearning strategies on CIFAR-10 with ResNet18 as the discriminator, the noise generator $G(\cdot)$ is trained.

The resulting unlearning degree is illustrated in Fig.\ref{unlearning degree}. The uniform random output accuracy for CIFAR-10 is around 10\%, therefore, when we conduct experiments with CIFAR-10, the range of unlearning degree is [0, 90\%]. The degree of unlearning indicates, beyond the model generalization caused high $Acc_{UL}$, the extent to which the machine unlearning actually eliminates the influence of unlearning data on the model. From the experimental results in Fig.\ref{unlearning degree}, it can be observed that Top-K achieved the best degree of unlearning (peaking at 88.36\% when 10\% data was unlearned), followed closely by Random-k, which reached a suboptimal level of unlearning (peaking at 86.04\% when 5\% of the data was unlearned). However, due to EU-K and CF-K freezing most of parameters and modifying the last $K$ layers, their degree of unlearning is significantly lower than that of Top-K and Random-k. Moreover, at lower ratio of unlearning data, such as 5\%, the degree of unlearning across different unlearning strategies if relatively higher. As unlearning ratio increases, the unlearning effect gradually diminishes. This implies that as more data needs to be forgotten, achieving better unlearning effects may become more difficult.

\section{Conclusion}\label{conclusion}

In this paper, we propose fine-grained model weights perturbation methods to achieve inexact machine unlearning. Our method ensures a higher degree of unlearning the information from dataset $D_{UL}$ while maintaining model indistinguishability and achieving largest acceleration effect. Furthermore, we design a novel machine unlearning degree quantification method using SPD-GAN to break the i.i.d property of the unlearning data. The results further demonstrate that our proposed Top-K based inexact unlearning strategy achieves the best unlearning effect and reveal that it is more difficult to unlearn more data.

\textbf{Limitations and Future Work} Although the Top-K strategy can achieve the best unlearning effect with the smallest perturbation ratio, the fastest acceleration and the optimal unlearning degree, our current analysis of parameter sensitivity in the model ignores the dependence between parameters. One possible direction is to redesign and select Top-K parameters for unlearning process considering the dependencies between parameters. By examining these dependencies during the unlearning process, it would become possible to interpret and explain the effects of unlearning on different components of the model, enhancing interpretability and transparency. This, in turn, is expected to facilitate better optimization of the unlearning process to improve overall model performance.

\section*{Acknowledgments}
The work is supported by the National Key Research and Development Program of China (2023YFB3002200), the National Natural Science Foundation of China (Grant Nos. 62225205, 92055213), the Science and Technology Program of Changsha (kh2301011), Shenzhen Basic Research Project (Natural Science Foundation) (JCYJ20210324140002006) and China Scholarship Council. Zhiwei Zuo did part of the work while she was a visiting student at NTU Singapore.

\ifCLASSOPTIONcaptionsoff
  \newpage
\fi

\bibliographystyle{IEEEtran}
\bibliography{main}

\begin{IEEEbiography}[{\includegraphics[width=1in,height=1.25in,clip,keepaspectratio]{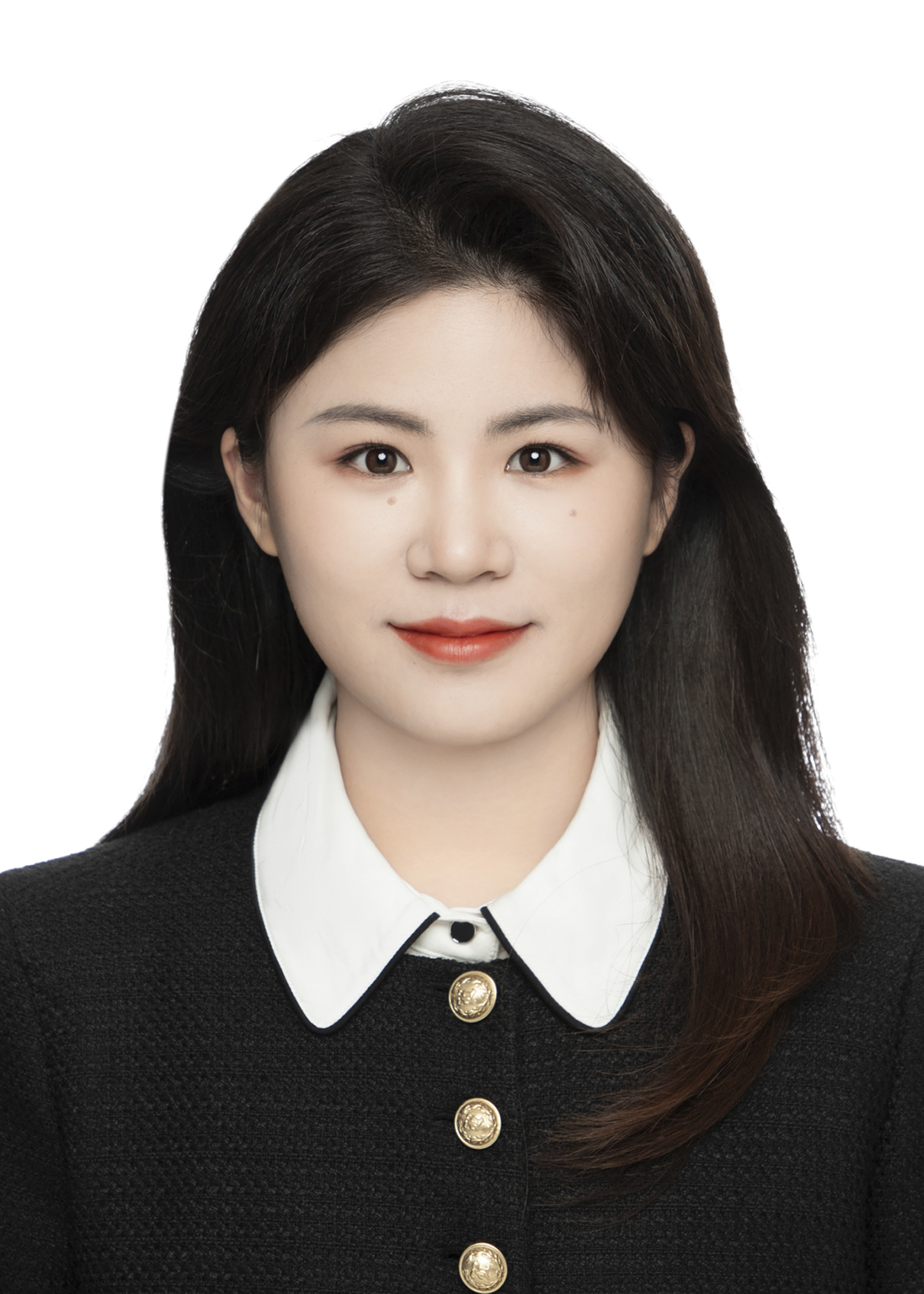}}]{Zhiwei Zuo}
is currently a Ph.D student with College of Computer Science and Electronic Engineering, Hunan University, China. In 2023, she was a visiting student at the School of Computer Science and Engineering in Nanyang Technological University (NTU), Singapore. Her research interests include cloud computing, machine learning, and machine unlearning.
\end{IEEEbiography}

\begin{IEEEbiography}[{\includegraphics[width=1in,height=1.25in,clip,keepaspectratio]{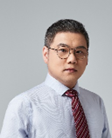}}]{Zhuo Tang} 
received the Ph.D. in computer science from Huazhong University of Science and Technology, China, in 2008. He is currently a professor of the College of Computer Science and Electronic Engineering at Hunan University. He is also the chief engineer of the National Supercomputing Center in Changsha. His majors are distributed computing system, cloud computing, and parallel processing for big data, including distributed machine learning, security model, parallel algorithms, and resources scheduling and management in these areas. He has published almost 120 journal articles and book chapters. He is a member of IEEE/ACM and CCF.
\end{IEEEbiography}

\begin{IEEEbiography}
[{\includegraphics[width=1in,height=1.25in,clip,keepaspectratio]{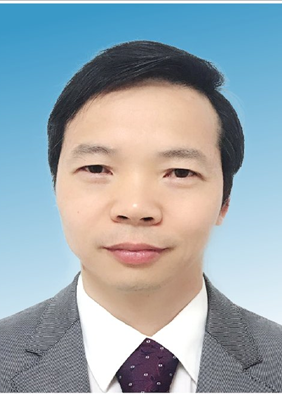}}]{Kenli Li}
received the PhD degree in computer science from Huazhong University of Science and Technology, China, in 2003. His major research areas include high performance computing, parallel computing, grid and cloud computing. He has published more than 300 research papers in international conferences and journals such as IEEE-TC, IEEE-TKDE, IEEE-TPDS, and ICDE. He is currently or has served on the editorial boards of the IEEE-TC, IEEE-TSUSC, and IEEE-TII. He is a senior member of IEEE and an outstanding member of CCF.
\end{IEEEbiography}

\begin{IEEEbiography}
[{\includegraphics[width=1in,height=1.25in,clip,keepaspectratio]{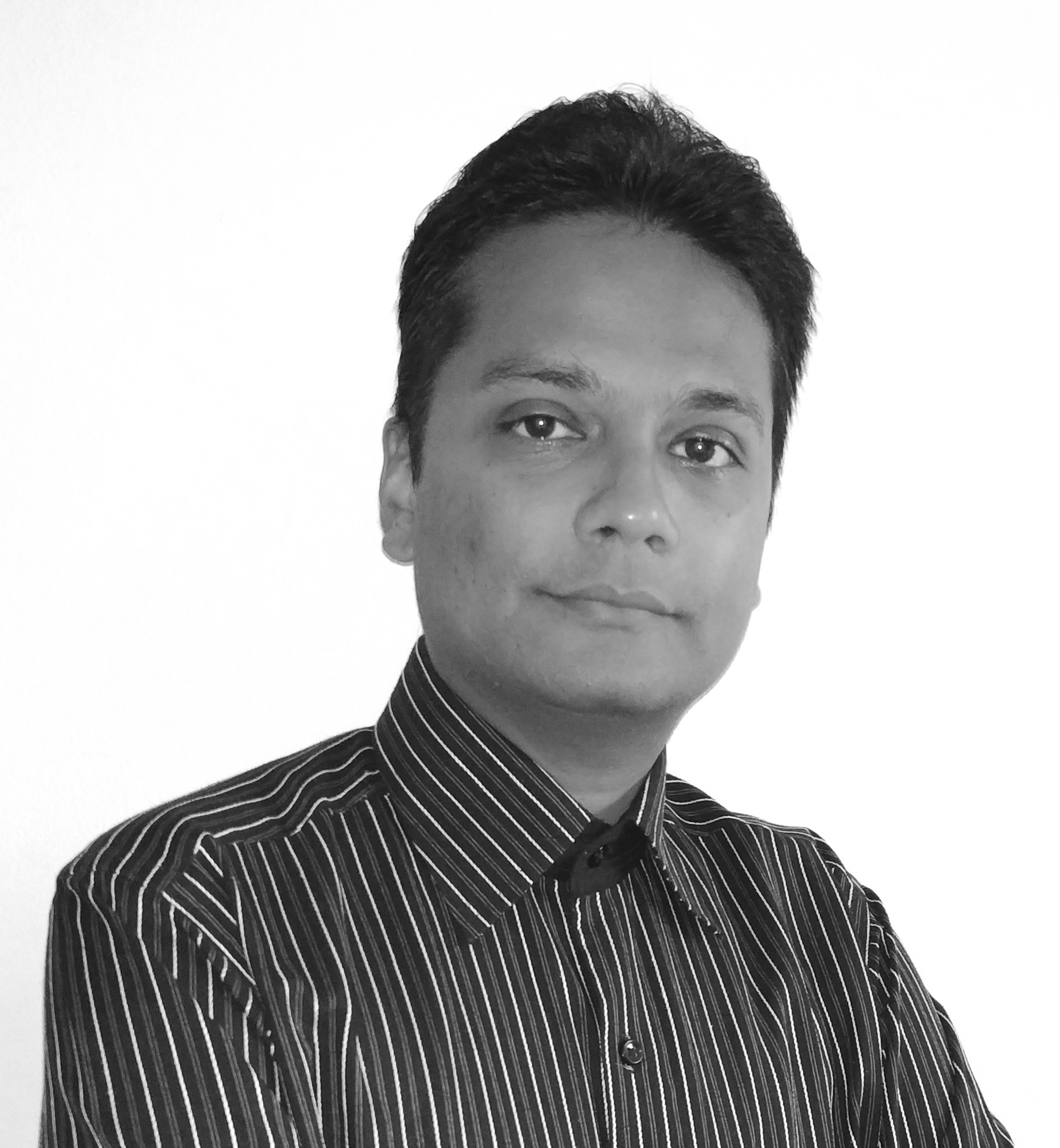}}]{Anwitaman Datta}
is a Professor of Cyber Security at De Montfort University, UK and is on leave from the College of Computing and Data Science,
Nanyang Technological University, Singapore where he holds an associate professorship.
His core research interests span the topics
of large-scale resilient distributed systems, information
security and applications of data analytics.
Presently, he is exploring topics at the intersection
of computer science, public policies \& regulations
along with the wider societal and (cyber)security
impact of technology. This includes the topics of
social media and network analysis, privacy, cyber-risk analysis and management,
cryptocurrency forensics, the governance of disruptive technologies,
as well as impact and use of disruptive technologies in digital societies and
in government’s technology stack.
\end{IEEEbiography}
\vfill

% that's all folks
\end{document}